\documentclass[11pt]{article}

\usepackage[final]{acl}

\usepackage{times}
\usepackage{latexsym}

\usepackage[T1]{fontenc}

\usepackage[utf8]{inputenc}

\usepackage{microtype}

\usepackage{inconsolata}
\usepackage{xspace}
\usepackage{graphicx}
\usepackage{tcolorbox}
\usepackage{xparse}

\usepackage{xcolor}
\usepackage{colortbl}
\usepackage{array}
\definecolor{oursbg}{RGB}{232,239,248}

\usepackage{bbding}
\usepackage{subfigure}
\usepackage[boxed,ruled,lined]{algorithm2e}
\usepackage{multirow}
\usepackage{makecell}

\usepackage{arydshln} 
\usepackage{enumerate}
\usepackage{enumitem}
\usepackage{bbm}
\usepackage{amsmath}
\usepackage{amssymb}
\usepackage{booktabs}
\usepackage{balance}
\usepackage{xcolor}
\usepackage{float}

\usepackage{graphicx}
\setkeys{Gin}{draft=false}

\usepackage{tcolorbox}
\tcbuselibrary{skins,breakable}
\usepackage{xparse}

\definecolor{planorange}{HTML}{B36A2E}
\definecolor{planbgorange}{HTML}{FBF1E6}

\definecolor{genblue}{HTML}{3F6F9F}
\definecolor{genbgblue}{HTML}{EEF4FA}

\definecolor{reviewred}{HTML}{A65353}
\definecolor{reviewbgred}{HTML}{FAEFEF}

\NewDocumentCommand{\promptboxbase}{m m m +m}{%
\begin{tcolorbox}[
    enhanced,
    breakable,
    width=\linewidth,
    colframe=#1,
    colback=#2,
    colbacktitle=#1,
    coltitle=white,
    title={#3},
    fonttitle=\bfseries\small,
    fontupper=\footnotesize\itshape,
    boxrule=0.7pt,
    arc=0mm,
    outer arc=0mm,
    boxsep=0pt,
    left=7pt,
    right=7pt,
    top=7pt,
    bottom=7pt,
    lefttitle=7pt,
    righttitle=7pt,
    toptitle=5pt,
    bottomtitle=5pt,
    before skip=7pt,
    after skip=9pt,
    before upper={
        \raggedright
        \sloppy
        \setlength{\parindent}{0pt}
    }
]
#4
\end{tcolorbox}%
}

\NewDocumentCommand{\planboxsimple}
{O{Planning/fact-cleaning prompt $P_{\mathrm{PLAN}}$}+m}{%
    \promptboxbase{planorange}{planbgorange}{#1}{#2}%
}

\NewDocumentCommand{\genboxsimple}
{O{Counterfactual generation prompt $P_{\mathrm{GEN}}$}+m}{%
    \promptboxbase{genblue}{genbgblue}{#1}{#2}%
}

\NewDocumentCommand{\reviewboxsimple}
{O{Reviewer prompt $P_{\mathrm{REV}}$}+m}{%
    \promptboxbase{reviewred}{reviewbgred}{#1}{#2}%
}

\newcommand{\ours}{PROSE\xspace}

\title{PROSE: Preservation-Constrained Selective Editing for Guilty Bias in Legal Judgment Prediction}

\author{
  \textbf{Yimeng Wang\textsuperscript{1,2}\thanks{Equal contribution}},
  \textbf{Zeyuan Tian\textsuperscript{1,*}},
  \textbf{Xinyu Wang\textsuperscript{3}},
  \textbf{Danyang Gao\textsuperscript{3}},
  \\
  \textbf{Chenglei Shen\textsuperscript{4}},
  \textbf{Zhongxiang Sun\textsuperscript{4}},
  \textbf{Lei La\textsuperscript{1}},
  \textbf{Weijie Yu\textsuperscript{1}\thanks{Corresponding author}},
  \\
  \textsuperscript{1}School of Artificial Intelligence and Data Science, University of International Business and Economics
  \\
\textsuperscript{2}Department of Computer Science, University of Chicago
\\
  \textsuperscript{3}School of Law, University of International Business and Economics
  \\
  \textsuperscript{4}Gaoling School of Artificial Intelligence, Renmin University of China
  \\\texttt{yu@uibe.edu.cn} 
\\
}

\begin{document}
\maketitle

\begin{abstract}
Legal judgment prediction models may retain a guilty verdict after a legally decisive condition has been altered such that the specified charge is no longer established.
We study this \emph{guilty bias} via selective repair over matched charge-established and charge-not-established cases, where the unsupported verdict should change while its supported counterpart remains stable. We construct CAIL-MC, a legally reviewed dataset of 11,478 matched pairs covering 261 Chinese criminal charges.
Held-out activation interventions show that contrasts learned from correctly
resolved pairs transfer to false-guilty cases, but target-reaching
interventions can also damage matched supported-guilty decisions. Motivated by
this observation,
we introduce \ours, which reconstructs
case-specific targets from a training-derived residual geometry, aligns them
with a local margin-descent signal, attenuates supported-guilty-sensitive
components, and realizes bounded targets through a preservation-regularized
low-rank parameter update. 
Across Qwen3-8B and GLM4-9B, \ours achieves stronger pairwise safe repair and evidence sensitivity, while preserving standard single-pass inference.

\end{abstract}

\section{Introduction}
\label{sec:intro}
Legal judgment prediction~(LJP)~\cite{zhong2018legal,zhang2023case,zhang2023contrastive,shi2025legalreasoner,han2026lawshift} asks whether case facts establish a
criminal charge. Beyond recognizing patterns in prior cases, an LJP system
should respond to legally decisive changes in the evidence. In particular,
when the available facts do not establish a charge, the principle of
\textit{in dubio pro reo}~\cite{raimondo2020indubio} requires doubt to be resolved in favor of the
accused. 

\begin{figure}[t]
    \centering
    \includegraphics[width=\columnwidth]{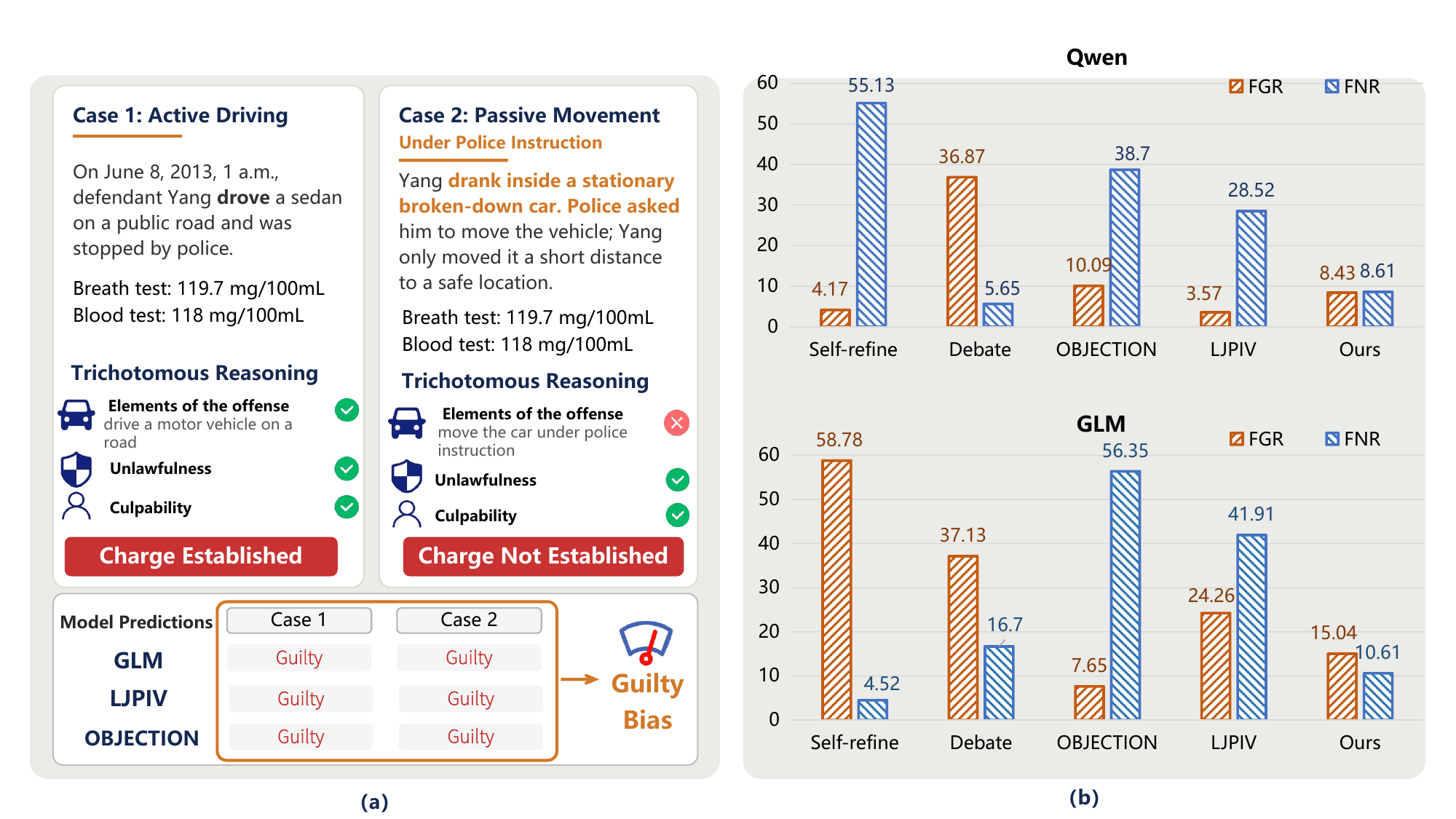}
    \caption{(a) Guilty bias and (b) its correction trade-off. 
    }
    \label{fig:intro}
\end{figure}

An LJP method may retain a guilty verdict
after the legal basis for that verdict has been altered. Figure~\ref{fig:intro}(a)
illustrates this behavior with two cases that share the specified charge and
surrounding context but differ in one decisive conduct condition. The evaluated
systems predict that the charge is established for both. We use
\emph{guilty bias} to denote this false-guilty tendency on legally matched
contrasts. The goal is therefore not to suppress guilty predictions globally,
but to correct the unsupported member while preserving its supported
counterpart.

Recent approaches reduce false-guilty predictions through additional non-guilty supervision~\cite{zhang2025beyond}, defense-oriented interaction~\cite{objection}, self-refinement~\cite{madaan2023selfrefine}, or
debate~\cite{chen2025debate}. Although they have achieved encouraging results, three challenges remain.
\textit{First}, aggregate error reduction does not establish evidence-sensitive correction. A lower false-guilty rate~(FGR) may result from a general downward shift in guilty scores, which can simultaneously increase the false-nonguilty rate~(FNR). As illustrated in
Figure~\ref{fig:intro}(b), reducing one error direction does not necessarily improve the other. Correction should therefore respond to the decisive legal
contrast rather than merely favor charge-not-established predictions.
\textit{Second}, repairability does not guarantee selective repair. Even an intervention
that moves a false-guilty case across the decision boundary may weaken the supported verdict of its matched charge-established counterpart or alter
unrelated correct predictions. A useful correction must therefore preserve both the paired counterpart and broader non-target behavior.
\textit{Third}, existing data-based correction requires multi-step training, whereas inference-time critique and debate repeat deliberation for every case. A practical correction should persist after a one-time update without requiring full retraining or repeated model calls at inference time.

To make these requirements measurable, we construct CAIL-MC, an LJP benchmark with two complementary tracks. The \textit{Pairwise Repair--Preservation Track}, derived from
CAIL2018~\cite{xiao2018cail}, contains 11,478
charge-established/charge-not-established pairs spanning 261 charges.
Guided by three-stage legal
reasoning~\cite{dubber2005theories,fletcher1978rethinking,zhang2025beyond},
each pair changes one legally decisive condition concerning offense
elements, unlawfulness, or culpability while keeping the surrounding facts
aligned, specifying both a repair target and its matched preservation control.
The \textit{Evidence-Sufficiency Generalization Track}, constructed from
LeCaRD~\cite{ma2021lecard} and Criminal-S~\cite{hu2018few}, comprises four
controlled conditions: full-supported, partial-sufficient,
partial-insufficient, and allegation-only.
It tests whether models retain supported charges and reject insufficiently
supported ones under corpus and construction shift, assessing evidence
sensitivity motivated by \textit{in dubio pro reo}.


We then test whether the base LLM contains an internal signal that can
support such a correction. Hidden-state contrasts estimated from correctly
resolved training pairs reduce the guilty margins of held-out false-guilty
cases. 
Yet interventions strong enough to
reach a charge-not-established target also reduce the margins of their matched
supported-guilty cases. 
This diagnosis exposes the central repair–preservation tension wherein errors are often internally repairable, but the available repair signal is not inherently selective.

Motivated by this finding, we introduce \ours, a 
preservation-constrained selective editing method. 
\ours addresses the repair--preservation tension in three stages. 
(1) it reconstructs each training target's matched residual within the geometry
learned from correctly resolved training pairs and defines a family of repair directions by interpolating the reconstruction with the negative gradient of the target verdict margin.
The residual component transfers structure from
successfully resolved legal contrasts, whereas the gradient provides a target-local signal for reducing the guilty margin. 
(2) \ours attenuates components
sensitive to supported-guilty behavior and searches within a case-adaptive cap for a small displacement that reaches a conservative negative margin.
This reduces overlap with protected behavior and limits unnecessary movement beyond the decision boundary.
(3) \ours realizes the feasible activation targets through preservation-regularized least squares over supported-guilty and non-target examples, followed by low-rank truncation and norm-bounded rescaling. Preservation is thus promoted during both target construction and parameter fitting. After this offline editing phase, the model performs ordinary single-pass inference.

Our contributions are threefold:
(1) We formulate guilty-bias mitigation as preservation-constrained selective
repair and introduce \ours, which realizes bounded activation targets
through a preservation-regularized parameter update.
(2) We construct CAIL-MC with two complementary tracks:
the Pairwise Repair--Preservation Track contains 11,478 reviewed matched
pairs across 261 charges, while the Evidence-Sufficiency Generalization
Track tests four evidence conditions from unseen corpora.
(3) Experiments on two backbones demonstrate stronger pairwise safe repair
than LJP and editing baselines, greater evidence sensitivity than SFT,
and less supported-verdict damage than a repair-matched uniform shift,
with lower offline cost than SFT and ordinary single-pass inference.

\section{Related Work}
\label{sec:related}

LJP maps case facts to judicial outcomes such as applicable law articles, charges, and sentences.
\citet{zhong2018legal} model dependencies among these prediction subtasks,
while \citet{hu2018few} address few-shot charge prediction through
discriminative legal attributes.
\citet{xu2020distinguish} focus on distinguishing confusing law articles, and
\citet{lyu2022improving} improve prediction through criminal-element
extraction.
More recent work incorporates legal precedents
\citep{wu2023precedent} and studies multi-task or multi-label prediction in
more realistic settings \citep{li2025legal}.

Recently, non-guilty judgment has received increasing attention.
\citet{zhang2025beyond} introduces charge-not-established outcomes and structures prediction
around offense elements, unlawfulness, and culpability.
\citet{objection} characterizes the tendency to over-predict guilt as
\emph{guilty bias} and uses an inference-time lawyer agent to challenge guilty
assumptions at each reasoning stage.
These approaches expand LJP beyond guilty-centered prediction, but do not
explicitly formulate correction as a matched requirement: changing an
unsupported verdict while preserving its minimally different supported
counterpart.

We address this gap by formulating guilty-bias mitigation as
\emph{preservation-constrained selective repair}. Each matched legal pair
jointly specifies a charge-not-established verdict to correct and a supported charge-established verdict to preserve. We realize these pairwise constraints through a persistent parameter update, connecting our approach to model editing~\citep{meng2022locating}. MEMIT~\citep{meng2023mass} performs scalable parameter updates to factual associations, while AlphaEdit~\citep{fang2024alphaedit} projects updates into a null space
associated with preserved knowledge. Unlike these factual editing methods, we modify an asymmetric
behavioral relation within each legal pair: the charge-not-established member
should change, while its charge-established counterpart and broader
non-target behavior should remain stable.

\section{Task and Benchmark}
\label{sec:task}
\subsection{Problem Formulation}
\label{sec:problem}

\paragraph{Charge-conditioned LJP.}
We study charge-conditioned verification. Given a case fact $x$ and a candidate
charge $c$, an LLM with parameters $\theta$ predicts whether $c$ is established. Let $z_\theta(1\mid x,c)$ and $z_\theta(0\mid x,c)$ denote the
answer-position logits for charge established and charge not established. We define
\begin{equation}
\small
\begin{aligned}
s_\theta(x,c)
&=z_\theta(1\mid x,c)-z_\theta(0\mid x,c),\\
\hat y_\theta(x,c)
&=\mathbbm{1}\!\left[s_\theta(x,c)>0\right].
\end{aligned}
\label{eq:verdict}
\end{equation}
We use the continuous margin for internal analysis and the binary prediction
for behavioral evaluation. 
Throughout, $\theta'$ denotes a candidate predictor evaluated relative to the fixed
reference model $\theta_0$.

\paragraph{Matched legal contrasts.}
An isolated prediction cannot distinguish evidence-sensitive behavior from a
general verdict preference. We therefore use a matched contrast
\begin{equation}
\small
\begin{aligned}
p_i&=(x_i^{\mathrm G},x_i^{\mathrm{NG}},c_i,d_i),~ d_i\in\mathcal D,~(y_i^{\mathrm G},y_i^{\mathrm{NG}})=(1,0),\\
 \mathcal D&=\{\text{\textsc{Offense}},\text{\textsc{Unlawfulness}},\text{\textsc{Culpability}}\}.
\end{aligned}
\label{eq:matched_pair}
\end{equation}
Both members concern the same candidate charge $c_i$ and are derived from a
common source case. The charge-not-established
member
$x_i^{\mathrm{NG}}$ changes one decisive condition at legal stage $d_i$ while
preserving the surrounding facts as closely as possible. Thus, $x_i^{\mathrm{NG}}$ specifies what
should change, whereas $x_i^{\mathrm G}$ provides the closest counterpart to preserve. 
The construction details are described in Sec.~\ref{sec:data}.

\paragraph{Repair and preservation.}
Let $\mathcal I^S$ denote the pairs in split $S$. The two side-level errors are
\begin{equation}
\small
\begin{aligned}
\mathrm{FGR}_S(\theta)
&=\Pr_{i\sim\mathcal I^S}
\!\left[\hat y_\theta(x_i^{\mathrm{NG}},c_i)=1\right],\\
\mathrm{FNR}_S(\theta)
&=\Pr_{i\sim\mathcal I^S}
\!\left[\hat y_\theta(x_i^{\mathrm G},c_i)=0\right].
\end{aligned}
\label{eq:fgr_fnr}
\end{equation}
Reporting both reveals whether reducing false-guilty errors merely shifts the
model toward charge-not-established predictions. These rates nevertheless
discard the correspondence within each pair. We therefore define the strict
false-guilty set under the fixed base model:
\begin{equation}
\small
\mathcal E_{\mathrm{sFG}}^S(\theta_0)
=
\left\{
i\in\mathcal I^S:
\left(
\hat y_{\theta_0}(x_i^{\mathrm G},c_i),
\hat y_{\theta_0}(x_i^{\mathrm{NG}},c_i)
\right)=(1,1)
\right\}.
\label{eq:strict_false_guilty}
\end{equation}
Its gold state is $(1,0)$ in the fixed order
$(x_i^{\mathrm G},x_i^{\mathrm{NG}})$. Repair concerns changing the unsupported
$x_i^{\mathrm{NG}}$ verdict, preservation concerns retaining the matched
$x_i^{\mathrm G}$ verdict and other correct behavior. 
For each reference model $\theta_0$ and split $S$, $\mathcal E_{\mathrm{sFG}}^S(\theta_0)$ is defined once under $\theta_0$ and held fixed across subsequent comparisons, keeping the evaluation population unchanged.
\subsection{CAIL-MC}
\label{sec:data}
CAIL-MC comprises two complementary tracks under the same
charge-conditioned verification task. The \emph{Pairwise
Repair--Preservation Track} (track~1) evaluates whether a model corrects an unsupported
verdict while preserving its closest supported counterpart. The
\emph{Evidence-Sufficiency Generalization Track} (track~2) evaluates whether this
behavior transfers when both the source corpus and the construction of
evidential support change. 

\paragraph{Pairwise Repair--Preservation Track.}
Based on CAIL2018~\citep{xiao2018cail}, we construct
11,478 charge-established/charge-not-established pairs spanning 261 Chinese
criminal charges. Following the three-stage legal reasoning perspective~\citep{zhang2025beyond},
each pair changes one decisive condition concerning offense elements,
unlawfulness, or culpability while preserving the surrounding facts as
closely as possible. The charge-not-established member specifies the verdict
that should change, whereas its matched charge-established member provides
the closest preservation control. The legal-stage annotation organizes
contrast construction and the stage-specific statistics used later; it is
not an additional prediction target. Construction includes outcome-expression
removal, legal-condition planning, counterfactual generation, automated
factual and legal checks, and stratified inspection by criminal-law
annotators. 
Variants derived from the same source case are assigned to the
same split, with the stage distribution preserved across splits.
The construction details are provided in App.~\ref{app:pairwise-track}.

\paragraph{Evidence-Sufficiency Generalization Track.}
To evaluate behavior beyond the matched construction, we derive four
evidence conditions from LeCaRD~\cite{ma2021lecard} and Criminal-S~\cite{hu2018few} cases.
The \emph{full-supported} condition retains the complete facts supporting the
candidate charge, while \emph{partial-sufficient} removes non-essential
information but retains sufficient support. In contrast,
\emph{partial-insufficient} preserves some charge-related facts but withholds
decisive support, and \emph{allegation-only} retains the accusation without
supporting factual evidence. The first two conditions are labeled charge
established, and the latter two charge not established. In particular we adopt partial-sufficient controls for shortcuts rooted in input length or retained information volume because a shorter description need not imply insufficient support. This track therefore tests whether a model retains supported charges
but rejects them when the presented evidence is insufficient, operationalizing
one measurable consequence of \textit{in dubio pro reo} under joint corpus
and evidence-construction shift. The construction details are provided in App.~\ref{app:evidence-track}.

\section{Diagnostic Analysis}
\label{sec:diagnosis}

We use temporary hidden-state interventions~\citep{li2023inference,zou2023representation} to test whether matched legal
contrasts provide a transferable repair signal and whether that signal is
selective. Contrast statistics are estimated from the training split, while
strict false-guilty pairs from the development split are held out for
diagnosis; the reference model remains frozen throughout.

\subsection{Transfer of Hidden-state Contrasts}
\label{sec:diag_layerwise}

Let $h_{\theta_0,t}^{\ell}(x,c)$ be the hidden state at prompt position $t$
of layer $\ell$, and let $\mathcal T$ denote the three positions immediately
preceding the answer\footnote{The three positions correspond to the tokens used by the
answer phrase ``the answer is'' in the English rendering of the prompt. We use the same
positions and intervention strength across layers and legal stages.}. 
For matched pair $i$, we define its
charge-not-established-minus-charge-established contrast as
\begin{equation}
\small
q_i^{\ell}
=\frac{1}{|\mathcal T|}\sum_{t\in\mathcal T}
\left[
h_{\theta_0,t}^{\ell}(x_i^{\mathrm{NG}},c_i)
-h_{\theta_0,t}^{\ell}(x_i^{\mathrm G},c_i)
\right].
\label{eq:diag_pair_direction}
\end{equation}
At each layer and legal stage, we average and normalize these contrasts over
training pairs whose two members are correctly predicted. We add the resulting
direction to the charge-not-established member of held-out strict false-guilty
pairs and measure NG-margin reduction: the original guilty-minus-charge-not-
established margin minus its intervened value. A positive reduction moves the
case toward charge not established. Forward directions are compared with
their reversals and norm-matched random directions.

\begin{figure}[t]
\centering
\includegraphics[width=0.9\columnwidth]{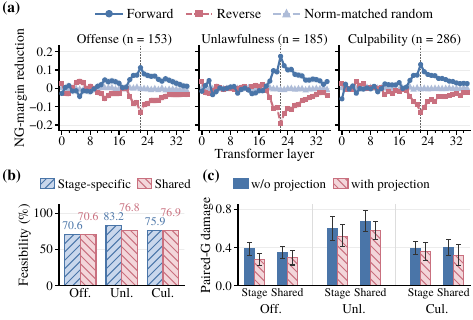}
\caption{Qwen3-8B diagnosis. (a) layer-wise NG-margin reduction under forward and control directions, (b) repair feasibility, and (c) paired-G
margin damage.
}
\label{fig:diag_qwen}
\end{figure}

Fig.~\ref{fig:diag_qwen} (a) shows that the forward direction reduces the
NG guilty margin in a localized middle-layer region across all three legal
stages, with the clearest response near Layer~22. Reversing the direction
generally reverses the effect, whereas random perturbations are smaller and
less stable. Contrasts learned from correctly resolved training pairs thus
transfer to held-out strict false-guilty cases.

\subsection{Repair Feasibility and Preservation}
\label{sec:diag_feasibility}

Directional transfer does not establish that the target verdict is reachable
or that reaching it preserves the matched supported decision. We therefore
reconstruct a case-specific direction using either stage-specific or shared
affine contrast geometry and vary its strength over a bounded grid. A case is
feasible if a tested strength moves its NG margin just below zero by a small
buffer, which marks a target beyond the decision boundary rather than a
calibrated confidence threshold. Fig.~\ref{fig:diag_qwen} (b)
shows feasibility of $70.6$--$83.2\%$ for stage-specific directions and
$70.6$--$76.9\%$ for shared directions. The transferred signal can therefore
reach the desired verdict for a substantial fraction of false-guilty cases.

We then apply each target-reaching displacement to its matched
charge-established member. Paired-G margin damage is the supported member's
pre-intervention minus post-intervention margin; positive values indicate a
weaker supported verdict. To reduce such damage, we remove the direction's
component in a subspace of verdict-margin gradients from correctly predicted
charge-established training cases. On pairs feasible both with and without
this preservation projection, damage decreases in all six stage/geometry
comparisons, from $0.349$--$0.674$ to $0.274$--$0.584$
(Fig.~\ref{fig:diag_qwen}c).

These results separate repairability from selectivity: matched contrasts
provide sufficient movement to repair false-guilty decisions, but that movement
can weaken their supported counterparts unless preservation is explicitly
controlled. 
Diagnostic details and GLM4-9B results, which yield consistent qualitative findings, are elaborated in App.~\ref{app:diagnosis}.

\section{Our Method: \ours}
\label{sec:method}

Diagnosis above shows that contrasts from correctly resolved pairs can move strict
false-guilty cases toward charge not established, but a repair-effective
displacement can also weaken the matched supported verdict. \ours addresses
this repair--preservation trade-off in two stages. (1) it constructs a bounded,
preservation-aware hidden-state target for each feasible training pair, and (2)
realizes these targets through an offline update constrained on supported and
non-target behavior.
We describe one stage-$d$ update at layer $\ell$ under the current model
$\vartheta$, initialized as $\theta_0$. Eligibility and preservation labels
are fixed under $\theta_0$, and representations are recomputed before each
stage update.

\subsection{Bounded Preservation-Aware Targets}
\label{sec:repair_targets}

For $i\in\mathcal E_{\mathrm{sFG}}^{\mathrm{train}}(\theta_0)$ with $d_i=d$, we
recompute $q_i^\ell$ under $\vartheta$ using
Eq.~\ref{eq:diag_pair_direction}. Correctly
resolved training contrasts at the same stage define a mean $\bar q_d^\ell$
and a rank-$r_q$ orthonormal basis $U_d^\ell$ from centered SVD. We reconstruct the pair contrast
in this affine geometry:
\begin{equation}
\small
\widehat q_i^\ell=
\bar q_d^\ell+
U_d^\ell(U_d^\ell)^\top(q_i^\ell-\bar q_d^\ell),
\label{eq:presume_affine}
\end{equation}
where the mean represents the average successful contrast, while the basis
retains the leading case-dependent deviations around it. Thus,
$\widehat q_i^\ell$ supplies a structured, case-conditioned candidate.

This reconstruction does not ensure
that it lowers the current target's verdict margin.
Let
$\widetilde{s}_{\vartheta}(x,c;\delta,\ell)$ denote the margin after adding
$\delta$ at layer $\ell$ to the answer-adjacent positions used in diagnosis,
and let
$\operatorname{unit}(a)=a/\lVert a\rVert_2$ for a nonzero vector $a$. We
therefore combine it with the target-local direction of first-order margin
descent:
\begin{equation}
\small
\begin{aligned}
g_i^\ell&=-\operatorname{unit}\!\left(
\left.\nabla_\delta\widetilde{s}_{\vartheta}
(x_i^{\mathrm{NG}},c_i;\delta,\ell)
\right|_{\delta=0}\right),\\
v_i^\ell(\lambda)&=\operatorname{unit}\!\left[
(1-\lambda)\operatorname{unit}(\widehat q_i^\ell)+\lambda g_i^\ell
\right].
\end{aligned}
\label{eq:presume_direction}
\end{equation}
Here, $0\leq\lambda\leq1$ balances the training-derived contrast and local
descent signal, which can improve alignment with the current decision boundary.

Repair directions may also affect supported-guilty behavior. Let $P_d^\ell$
be a rank-$r_p$ orthonormal basis of leading margin-gradient directions from correctly
predicted charge-established training cases at stage $d$. We attenuate the
overlapping component:
\begin{equation}
\small
u_i^\ell=\operatorname{unit}\!\left[
v_i^\ell(\lambda)-\gamma P_d^\ell(P_d^\ell)^\top v_i^\ell(\lambda)
\right].
\label{eq:presume_attenuation}
\end{equation}
With $\gamma=1/(1+\rho)$ and $\rho\geq0$, $\rho=0$ fully removes this
component, whereas $\rho>0$ retains some repair-relevant overlap. This soft
attenuation reduces sensitivity to supported verdicts without discarding all
overlapping repair information.

The direction specifies where to move a target, not how far. Let $\beta\geq0$
be the displacement magnitude. A global cap $\beta_{\max}>0$ and a
case-relative scale $\kappa>0$ bound the search by
\begin{equation}
\small
B_i=\min\!\left\{\beta_{\max},\kappa\|\widehat q_i^\ell\|_2\right\}.
\label{eq:presume_beta_cap}
\end{equation}
We search for a small strength $\beta_i^\star$ reaching a negative margin
buffer $\tau<0$, and define the activation target $t_i$:
\begin{equation}
\small
\widetilde{s}_{\vartheta}(x_i^{\mathrm{NG}},c_i;
\beta_i^\star u_i^\ell,\ell)\leq\tau,
\label{eq:presume_beta}
\end{equation}
where $t_i=\beta_i^\star u_i^\ell, 0\leq\beta_i^\star\leq B_i$. 
The bounded, case-specific search seeks only the displacement needed for the
chosen direction, thereby limiting avoidable target movement. 
Let $\mathcal F_d$ contain the pairs
for which the search finds such a strength and only their displacements $t_i$
enter parameter fitting.

\subsection{Preservation-Constrained Update}
\label{sec:parameter_update}
Since the targets above are temporary, to make their effects persist without paired
inputs or per-case search at inference time, we fit an additive update
$\Delta W$ to $W$, block notation for the selected projection weights,
for stage $d$. Let $K_{\mathcal F}$ and
$T_{\mathcal F}$ collect the module-input keys and target displacements from
$\mathcal F_d$, with one column per intervention position and
$n_{\mathcal F}$ columns in total. Each position receives its pair's $t_i$.
We retain three base-correct
zero-change pools: supported charge-established cases ($\mathrm G$), charge-not-established members of correctly resolved
same-stage pairs ($\mathrm O$),
and correct cases from other stages ($\mathrm X$). For pool $a$, $K_a$, $n_a$,
and $\omega_a\geq0$ denote its key matrix, column count, and penalty weight.
We minimize
\begin{equation}
\small
\begin{aligned}
\mathcal L(\Delta W)={}&
\frac{\|\Delta W K_{\mathcal F}-T_{\mathcal F}\|_F^2}{n_{\mathcal F}}
+\mu\|\Delta W\|_F^2\\
&+\sum_{a\in\{\mathrm G,\mathrm O,\mathrm X\}}
\frac{\omega_a}{n_a}\|\Delta W K_a\|_F^2,
\qquad \mu>0.
\end{aligned}
\label{eq:presume_update_objective}
\end{equation}
The first term fits feasible repair targets, the remaining terms penalize changes to supported, same-stage non-target, and cross-stage behavior, while
$\mu$ regularizes update magnitude. For fixed targets and keys, $\mu>0$
yields a unique closed-form ridge solution, without iterative optimization
of model weights. We obtain a rank-$r_W$ approximation of this solution
and rescale it to satisfy a relative
Frobenius-norm budget $\varepsilon>0$ before setting $W'=W+\Delta W$.
These subsequent controls limit update capacity and magnitude; they do not
solve the jointly rank- and norm-constrained objective exactly.
After the stage updates, $\theta'$ performs ordinary
single-pass inference. The derivation, search, and execution details are
provided in App.~\ref{app:method}.

\section{Experiments}
\label{sec:experiments}
We conduct experiments to address the following research questions:
\textbf{RQ1:} Does \ours reduce false-guilty verdicts without trading them for false non-guilty ones, on matched pairs and on evidence-graded cases from unseen corpora?
\textbf{RQ2:} Does \ours repair the unsupported member while preserving its supported counterpart, rather than shifting all verdicts?
\textbf{RQ3:} Which preservation components make the repair selective?
\textbf{RQ4:} What does the correction cost, offline and at inference?
\subsection{Experimental Setup}

\label{sec:setup}
\paragraph{Datasets.}
We use \textit{Pairwise Repair--Preservation Track} for in-domain training and matched-pair evaluation,
\textit{Evidence-Sufficiency Generalization Track} for evidence-controlled cross-corpus evaluation. 
Dataset statistics are summarized in
Tab.~\ref{tab:data}, the construction and annotation are
provided in App.~\ref{app:dataset}.


\begin{table}[t]
\centering
\small
\setlength{\tabcolsep}{3pt}
\begin{tabular*}{\columnwidth}{
    @{\extracolsep{\fill}}llrrr@{}
}
\toprule
Track & Source & Train & Dev & Test \\
\midrule
Pairwise & CAIL2018 & 9,181 & 1,147 & 1,150 \\
\multicolumn{5}{@{}c@{}}{
    \footnotesize Stage distribution:
    O 35.7\%, U 17.8\%, C 46.5\%
} \\
\midrule
Evidence & LeCaRD     & -- & -- & 280 \\
         & Criminal-S & -- & -- & 800 \\
\bottomrule
\end{tabular*}
\caption{CAIL-MC statistics. Pairwise counts are matched pairs;
evidence counts are variants, four per source case
(70 LeCaRD; 200 Criminal-S).
O/U/C denote Offense/Unlawfulness/Culpability.}
\label{tab:data}
\end{table}

\begin{table*}[t]
\centering
\small
\setlength{\tabcolsep}{2.1pt}
\renewcommand{\arraystretch}{1.05}
\resizebox{\textwidth}{!}{%
\begin{tabular}{l ccccccc cccc cccc}
\toprule
& \multicolumn{7}{c}{\textbf{CAIL2018 (Pairwise Repair–Preservation)}} & \multicolumn{4}{c}{\textbf{LeCaRD (Evidence-Sufficiency Generalization)}} & \multicolumn{4}{c}{\textbf{Criminal-S (Evidence-Sufficiency Generalization)}} \\
\cmidrule(lr){2-8}\cmidrule(lr){9-12}\cmidrule(lr){13-16}
& \multicolumn{4}{c}{Case level} & \multicolumn{3}{c}{Pair level} & \multicolumn{4}{c}{Case level} & \multicolumn{4}{c}{Case level} \\
\cmidrule(lr){2-5}\cmidrule(lr){6-8}\cmidrule(lr){9-12}\cmidrule(lr){13-16}
\textbf{Method} & Acc$\uparrow$ & F1$\uparrow$ & FGR$\downarrow$ & FNR$\downarrow$ & Rep.$\uparrow$ & G-Ret.$\uparrow$ & PSR$\uparrow$ & Acc$\uparrow$ & F1$\uparrow$ & FGR$\downarrow$ & FNR$\downarrow$ & Acc$\uparrow$ & F1$\uparrow$ & FGR$\downarrow$ & FNR$\downarrow$ \\
\midrule
\multicolumn{16}{c}{\emph{Qwen3-8B-Base}} \\
Base & 72.65{\scriptsize$\pm$1.43} & 70.58{\scriptsize$\pm$1.77} & 53.91{\scriptsize$\pm$2.82} & 0.78{\scriptsize$\pm$0.50} & -- & -- & -- & 55.71{\scriptsize$\pm$2.99} & 46.91{\scriptsize$\pm$5.22} & 85.00{\scriptsize$\pm$6.59} & 3.57{\scriptsize$\pm$3.67} & 66.88{\scriptsize$\pm$2.58} & 63.63{\scriptsize$\pm$3.36} & 63.00{\scriptsize$\pm$5.05} & 3.25{\scriptsize$\pm$2.00} \\
SFT & \textbf{98.87}{\scriptsize$\pm$0.44} & \textbf{98.87}{\scriptsize$\pm$0.44} & \textbf{2.00}{\scriptsize$\pm$0.83} & \textbf{0.26}{\scriptsize$\pm$0.30} & \textbf{96.27}{\scriptsize$\pm$1.53} & \textbf{100.00}{\scriptsize$\pm$0.00} & \textbf{96.27}{\scriptsize$\pm$1.53} & 49.64{\scriptsize$\pm$2.71} & 37.60{\scriptsize$\pm$3.45} & 94.29{\scriptsize$\pm$3.79} & \textbf{6.43}{\scriptsize$\pm$5.26} & 51.00{\scriptsize$\pm$1.12} & 36.72{\scriptsize$\pm$1.83} & 96.50{\scriptsize$\pm$1.78} & \textbf{1.50}{\scriptsize$\pm$1.58} \\
\cmidrule(lr){1-16}
Self-Refine & 70.35{\scriptsize$\pm$1.49} & 68.29{\scriptsize$\pm$1.79} & 4.17{\scriptsize$\pm$1.13} & 55.13{\scriptsize$\pm$2.77} & \underline{95.79}{\scriptsize$\pm$1.63} & 39.71{\scriptsize$\pm$3.66} & 36.79{\scriptsize$\pm$3.65} & 54.29{\scriptsize$\pm$4.56} & 47.90{\scriptsize$\pm$6.33} & \textbf{10.71}{\scriptsize$\pm$5.13} & 80.71{\scriptsize$\pm$7.98} & 54.88{\scriptsize$\pm$1.70} & 44.86{\scriptsize$\pm$2.97} & \textbf{2.50}{\scriptsize$\pm$1.53} & 87.75{\scriptsize$\pm$3.37} \\
OBJECTION & 75.61{\scriptsize$\pm$1.59} & 75.10{\scriptsize$\pm$1.68} & 10.09{\scriptsize$\pm$1.81} & 38.70{\scriptsize$\pm$2.82} & 89.14{\scriptsize$\pm$2.56} & 61.59{\scriptsize$\pm$3.83} & 53.81{\scriptsize$\pm$3.91} & \textbf{73.57}{\scriptsize$\pm$4.75} & \textbf{73.22}{\scriptsize$\pm$5.03} & \underline{15.00}{\scriptsize$\pm$5.36} & 37.86{\scriptsize$\pm$9.23} & \textbf{73.38}{\scriptsize$\pm$3.00} & \textbf{73.27}{\scriptsize$\pm$3.05} & 20.50{\scriptsize$\pm$4.03} & 32.75{\scriptsize$\pm$5.14} \\
Debate & 78.74{\scriptsize$\pm$1.45} & 78.21{\scriptsize$\pm$1.56} & 36.87{\scriptsize$\pm$2.72} & 5.65{\scriptsize$\pm$1.32} & 44.08{\scriptsize$\pm$3.79} & 96.27{\scriptsize$\pm$1.48} & 41.49{\scriptsize$\pm$3.71} & 66.79{\scriptsize$\pm$4.60} & 65.35{\scriptsize$\pm$5.11} & 53.57{\scriptsize$\pm$8.40} & \underline{12.86}{\scriptsize$\pm$6.51} & 67.88{\scriptsize$\pm$3.08} & \underline{67.40}{\scriptsize$\pm$3.19} & 44.25{\scriptsize$\pm$4.91} & 20.00{\scriptsize$\pm$4.07} \\
LJPIV & 83.96{\scriptsize$\pm$1.45} & 83.70{\scriptsize$\pm$1.52} & \underline{3.57}{\scriptsize$\pm$1.09} & 28.52{\scriptsize$\pm$2.65} & 93.52{\scriptsize$\pm$1.96} & 69.53{\scriptsize$\pm$3.57} & 64.67{\scriptsize$\pm$3.80} & 61.79{\scriptsize$\pm$4.69} & 60.37{\scriptsize$\pm$5.54} & 19.29{\scriptsize$\pm$6.45} & 57.14{\scriptsize$\pm$10.48} & 68.25{\scriptsize$\pm$2.62} & 66.48{\scriptsize$\pm$3.24} & \underline{8.75}{\scriptsize$\pm$2.60} & 54.75{\scriptsize$\pm$5.89} \\
\cmidrule(lr){1-16}
MEMIT & 51.22{\scriptsize$\pm$1.83} & 49.98{\scriptsize$\pm$1.92} & 64.52{\scriptsize$\pm$2.87} & 33.04{\scriptsize$\pm$2.74} & 34.68{\scriptsize$\pm$3.92} & 68.72{\scriptsize$\pm$3.68} & 19.94{\scriptsize$\pm$3.29} & 37.86{\scriptsize$\pm$4.40} & 35.98{\scriptsize$\pm$4.02} & 79.29{\scriptsize$\pm$5.74} & 45.00{\scriptsize$\pm$9.43} & 36.38{\scriptsize$\pm$2.45} & 34.39{\scriptsize$\pm$2.60} & 46.25{\scriptsize$\pm$4.75} & 81.00{\scriptsize$\pm$4.36} \\
AlphaEdit & 62.78{\scriptsize$\pm$1.64} & 60.44{\scriptsize$\pm$1.93} & 61.57{\scriptsize$\pm$2.94} & 12.87{\scriptsize$\pm$1.94} & 50.41{\scriptsize$\pm$3.97} & 87.52{\scriptsize$\pm$2.57} & 41.98{\scriptsize$\pm$3.78} & 40.71{\scriptsize$\pm$4.34} & 36.52{\scriptsize$\pm$4.02} & 85.00{\scriptsize$\pm$5.47} & 33.57{\scriptsize$\pm$9.14} & 37.50{\scriptsize$\pm$2.78} & 36.01{\scriptsize$\pm$2.55} & 77.75{\scriptsize$\pm$3.46} & 47.25{\scriptsize$\pm$5.75} \\
\midrule
\rowcolor{oursbg} \textbf{PROSE} & \underline{91.48}{\scriptsize$\pm$1.09} & \underline{91.48}{\scriptsize$\pm$1.09} & 8.43{\scriptsize$\pm$1.63} & 8.61{\scriptsize$\pm$1.57} & 85.25{\scriptsize$\pm$2.84} & 90.76{\scriptsize$\pm$2.17} & \underline{76.34}{\scriptsize$\pm$3.30} & \underline{70.36}{\scriptsize$\pm$4.07} & \underline{69.57}{\scriptsize$\pm$4.26} & 45.71{\scriptsize$\pm$7.59} & 13.57{\scriptsize$\pm$7.29} & \underline{68.38}{\scriptsize$\pm$2.67} & 67.33{\scriptsize$\pm$2.87} & 49.50{\scriptsize$\pm$4.80} & \underline{13.75}{\scriptsize$\pm$3.99} \\
\midrule
\multicolumn{16}{c}{\emph{GLM-4-9B-Base}} \\
Base & 74.74{\scriptsize$\pm$1.47} & 73.63{\scriptsize$\pm$1.67} & 45.74{\scriptsize$\pm$2.90} & 4.78{\scriptsize$\pm$1.25} & -- & -- & -- & 57.50{\scriptsize$\pm$3.55} & 49.91{\scriptsize$\pm$5.39} & 81.43{\scriptsize$\pm$6.78} & 3.57{\scriptsize$\pm$4.27} & 57.75{\scriptsize$\pm$1.95} & 48.70{\scriptsize$\pm$3.38} & 84.25{\scriptsize$\pm$3.92} & 0.25{\scriptsize$\pm$0.49} \\

SFT & 
\textbf{98.78}{\scriptsize$\pm$0.46} &
\textbf{98.78}{\scriptsize$\pm$0.46} &
\textbf{2.26}{\scriptsize$\pm$0.91} &
\textbf{0.17}{\scriptsize$\pm$0.22} &
\textbf{94.98}{\scriptsize$\pm$1.92} &
\textbf{99.61}{\scriptsize$\pm$0.48} &
\textbf{94.59}{\scriptsize$\pm$2.03} &
50.36{\scriptsize$\pm$1.79} &
36.98{\scriptsize$\pm$3.54} &
95.71{\scriptsize$\pm$3.21} &
\underline{3.57}{\scriptsize$\pm$3.93} &
50.63{\scriptsize$\pm$1.00} &
35.94{\scriptsize$\pm$1.51} &
97.25{\scriptsize$\pm$1.63} &
\textbf{1.50}{\scriptsize$\pm$1.75}
\\

\cmidrule(lr){1-16}

Self-Refine &
68.35{\scriptsize$\pm$1.46} &
65.83{\scriptsize$\pm$1.80} &
58.78{\scriptsize$\pm$2.78} &
\underline{4.52}{\scriptsize$\pm$1.22} &
6.37{\scriptsize$\pm$2.09} &
\underline{98.07}{\scriptsize$\pm$1.18} &
5.98{\scriptsize$\pm$2.02} &
56.43{\scriptsize$\pm$3.11} &
47.40{\scriptsize$\pm$5.15} &
85.00{\scriptsize$\pm$5.97} &
\textbf{2.14}{\scriptsize$\pm$2.48} &
50.13{\scriptsize$\pm$0.24} &
33.61{\scriptsize$\pm$0.53} &
\textbf{0.00}{\scriptsize$\pm$0.00} &
99.75{\scriptsize$\pm$0.48}
\\


OBJECTION &
68.00{\scriptsize$\pm$1.51} &
65.98{\scriptsize$\pm$1.79} &
\underline{7.65}{\scriptsize$\pm$1.56} &
56.35{\scriptsize$\pm$2.84} &
88.22{\scriptsize$\pm$2.87} &
45.75{\scriptsize$\pm$4.25} &
36.68{\scriptsize$\pm$4.24} &
\underline{59.64}{\scriptsize$\pm$4.88} &
57.19{\scriptsize$\pm$5.83} &
\underline{16.43}{\scriptsize$\pm$5.84} &
64.29{\scriptsize$\pm$9.05} &
\underline{69.38}{\scriptsize$\pm$3.08} &
\underline{68.80}{\scriptsize$\pm$3.28} &
17.00{\scriptsize$\pm$3.43} &
44.25{\scriptsize$\pm$5.49}
\\

Debate &
73.09{\scriptsize$\pm$1.62} &
72.80{\scriptsize$\pm$1.67} &
37.13{\scriptsize$\pm$2.75} &
16.70{\scriptsize$\pm$2.14} &
38.22{\scriptsize$\pm$4.05} &
88.42{\scriptsize$\pm$2.75} &
33.59{\scriptsize$\pm$3.95} &
57.86{\scriptsize$\pm$5.41} &
54.69{\scriptsize$\pm$6.28} &
68.57{\scriptsize$\pm$8.32} &
15.71{\scriptsize$\pm$6.46} &
61.13{\scriptsize$\pm$2.59} &
57.32{\scriptsize$\pm$3.23} &
68.75{\scriptsize$\pm$4.45} &
9.00{\scriptsize$\pm$2.81}
\\

LJPIV &
66.91{\scriptsize$\pm$2.18} &
66.65{\scriptsize$\pm$2.22} &
24.26{\scriptsize$\pm$2.53} &
41.91{\scriptsize$\pm$2.90} &
67.37{\scriptsize$\pm$4.05} &
54.44{\scriptsize$\pm$4.35} &
42.66{\scriptsize$\pm$4.34} &
50.71{\scriptsize$\pm$2.60} &
39.16{\scriptsize$\pm$3.58} &
92.86{\scriptsize$\pm$3.93} &
5.71{\scriptsize$\pm$4.63} &
49.25{\scriptsize$\pm$0.98} &
33.64{\scriptsize$\pm$0.89} &
99.25{\scriptsize$\pm$0.80} &
\underline{2.25}{\scriptsize$\pm$1.95}
\\

\cmidrule(lr){1-16}

MEMIT &
54.57{\scriptsize$\pm$1.30} &
48.37{\scriptsize$\pm$1.74} &
10.78{\scriptsize$\pm$1.78} &
80.09{\scriptsize$\pm$2.25} &
89.58{\scriptsize$\pm$2.66} &
17.37{\scriptsize$\pm$3.11} &
12.36{\scriptsize$\pm$2.74} &
54.64{\scriptsize$\pm$3.47} &
48.46{\scriptsize$\pm$5.43} &
\textbf{10.71}{\scriptsize$\pm$4.83} &
80.00{\scriptsize$\pm$7.89} &
55.63{\scriptsize$\pm$1.78} &
48.53{\scriptsize$\pm$2.93} &
\underline{7.25}{\scriptsize$\pm$2.23} &
81.50{\scriptsize$\pm$4.00}
\\

AlphaEdit &
60.00{\scriptsize$\pm$1.62} &
57.86{\scriptsize$\pm$1.84} &
17.48{\scriptsize$\pm$2.23} &
62.52{\scriptsize$\pm$2.82} &
\underline{90.15}{\scriptsize$\pm$2.51} &
31.47{\scriptsize$\pm$3.99} &
24.71{\scriptsize$\pm$3.64} &
32.14{\scriptsize$\pm$3.89} &
31.72{\scriptsize$\pm$4.25} &
60.00{\scriptsize$\pm$4.79} &
75.71{\scriptsize$\pm$8.17} &
30.88{\scriptsize$\pm$2.24} &
30.71{\scriptsize$\pm$2.33} &
64.25{\scriptsize$\pm$3.14} &
74.00{\scriptsize$\pm$4.45}
\\

\midrule

\rowcolor{oursbg}
\textbf{PROSE} &
\underline{87.17}{\scriptsize$\pm$1.34} &
\underline{87.17}{\scriptsize$\pm$1.35} &
15.04{\scriptsize$\pm$2.07} &
10.61{\scriptsize$\pm$1.90} &
67.37{\scriptsize$\pm$4.06} &
94.02{\scriptsize$\pm$1.94} &
\underline{62.16}{\scriptsize$\pm$4.15} &
\textbf{66.79}{\scriptsize$\pm$5.06} &
\textbf{66.06}{\scriptsize$\pm$5.41} &
47.86{\scriptsize$\pm$9.86} &
18.57{\scriptsize$\pm$8.04} &
\textbf{71.88}{\scriptsize$\pm$2.33} &
\textbf{70.63}{\scriptsize$\pm$2.63} &
48.75{\scriptsize$\pm$4.63} &
7.50{\scriptsize$\pm$3.07}
\\
\bottomrule
\end{tabular}%
}
\caption{Results (\%) on both CAIL-MC tracks.
Entries are point estimates $\pm$ 1.96\% bootstrap standard errors
(1,000 resamples over pairs for repair--preservation and cases for
evidence-sufficiency evaluation).
Best and second-best results per backbone, excluding Base, are bold and underlined.
Base's pair-level scores are omitted:
Rep.\ = PSR = 0 and G-Ret.\ = 100 by construction.}
\label{tab:main}
\end{table*}

\paragraph{Baselines.}
We consider three groups of baselines.
 (1) Base denotes the raw LLM without corrective intervention, and SFT denotes LoRA fine-tuning~\cite{hu2022lora} across all layers using verdicts from both members of each pair.
(2) \textbf{LJP methods} include: LJPIV~\citep{zhang2025beyond}, Self-Refine~\citep{madaan2023selfrefine}, Debate-Feedback~\citep{chen2025debate},  and OBJECTION~\citep{objection}.
(3) \textbf{Model editing methods} include: MEMIT~\citep{meng2023mass} and AlphaEdit~\citep{fang2024alphaedit}.
Detailed descriptions are provided in App.~\ref{app:baselines}.

\paragraph{Evaluation and Implementation.}
We evaluate model behavior at both case and pair levels.
At the case level, in addition to accuracy and Macro-F1, we report FGR and FNR. These metrics are interpreted jointly to uncover one-sided verdict shifts, as defined in Eq.~\ref{eq:fgr_fnr}.
At the pair level, we evaluate the strict false-guilty set
$\mathcal E_{\mathrm{sFG}}^S(\theta_0)$, fixed by the reference model for each
backbone and evaluation set.
NG repair (Rep.) measures whether the charge-not-established member is
corrected, matched-G retention (G-Ret.) measures whether its supported
counterpart remains correct, and pairwise safe repair (PSR) requires both
conditions.
Formal definitions are provided in App.~\ref{app:pair_metrics}.
We employ Qwen3-8B-Base\footnote{\url{https://huggingface.co/Qwen/Qwen3-8B-Base}} and GLM-4-9B-Base\footnote{\url{https://huggingface.co/zai-org/glm-4-9b-hf}}. For these two models, \ours performs edits at Layer~22 and Layer~20, respectively (Sec.~\ref{sec:diagnosis}). Hyperparameters are tuned on the development split.
Implementation details are provided in App.~\ref{app:hyper}.

\subsection{Main Results (RQ1)}
\label{sec:main_results}
To evaluate guilty-bias reduction and selective repair, we compare methods on both CAIL-MC tracks.
From Tab.~\ref{tab:main}, we observe three patterns.
(1) \ours achieves the highest in-corpus PSR among LJP and editing baselines on both backbones, demonstrating stronger joint repair and matched preservation. On CAIL2018 track, \ours ranks second only to SFT and keeps both FNR and FGR at or below 8.61 (Qwen) and 15.04 (GLM), and its PSR is the highest after SFT.
(2) SFT achieves the strongest performance on matched pairs, yet this advantage does not carry over to evidence-sufficiency evaluation (Acc: 49.64–51.00), where false-guilty errors dominate (94.29–97.25).
(3) \ours achieves the best cross-corpus Macro-F1 on GLM and competitive performance on Qwen, avoiding the extreme one-sided error profiles exhibited by several baselines. For example, Self-Refine on Qwen and MEMIT on GLM obtain low FGR at the
expense of high FNR.
These results distinguish \ours's joint repair and preservation from error reduction through a general verdict shift.

\begin{figure}[t]
  \centering
  \includegraphics[width=0.9\columnwidth]{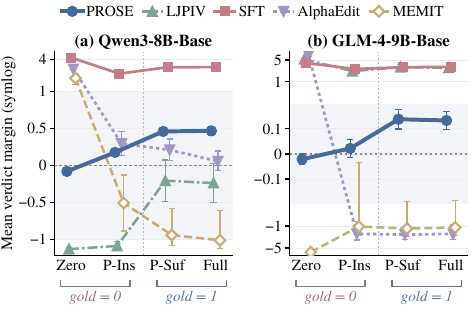}
  \caption{Mean verdict margin as evidence is removed from the 200 Criminal-S cases of Evidence Track, with (a) Qwen3-8B-Base and (b) GLM-4-9B-Base.}
  \label{fig:evidence}
\end{figure}
To further probe whether SFT's transfer gap is consistent with construction-specific
shortcut learning~\citep{gururangan2018annotation,geirhos2020shortcut}, we compare verdict margins
across four evidence conditions for the same 200 Criminal-S cases. As shown in Fig.~\ref{fig:evidence}, SFT retains strongly positive margins even without supporting evidence,
whereas other baselines exhibit different failures:
LJPIV on Qwen remains negative even with sufficient evidence, and MEMIT and AlphaEdit can assign higher margins to allegation-only inputs than to fully supported ones. 
In contrast, \ours retains positive margins with sufficient evidence,
moves closer to the decision boundary with insufficient evidence, and
becomes negative for allegation-only inputs.
Although partial-insufficient margins remain slightly positive, this
response better reflects the evidence-sensitive behavior motivated by
\textit{in dubio pro reo}, while preserving supported verdicts.

\subsection{Repair and Preservation Analysis (RQ2)}
\label{sec:selectivity}
\begin{figure}[t]
  \centering
  \includegraphics[width=0.9\columnwidth]{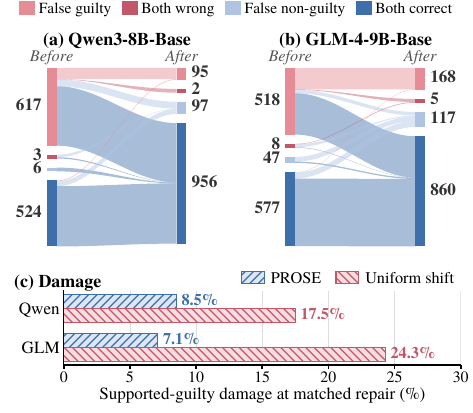}
  \caption{Repair and preservation. (a--b) Pair states before and after PROSE.
  (c) Supported-guilty damage at matched repair; lower is better.}
  \label{fig:selectivity}
\end{figure}
To investigate whether \ours repairs false-guilty decisions while preserving supported verdicts, we track the joint predictions of each test pair before and after correction (Fig~\ref{fig:selectivity}a--b). Most false-guilty pairs become jointly correct, and most originally correct pairs remain correct. Nevertheless, false-non-guilty pairs increase, showing that repair still incurs a preservation cost.

We further compare \ours with a uniform margin offset calibrated to match its repair rate on base-misclassified charge-not-established cases. Damage is measured as the fraction of base-correct charge-established cases that become incorrect. At matched repair, \ours causes substantially less damage: 8.5\% versus 17.5\% on Qwen and 7.1\% versus 24.3\% on GLM (Fig~\ref{fig:selectivity}c). Thus, \ours preserves supported verdicts better than a uniform shift toward charge-not-established predictions, demonstrating more selective repair.

Charge-level effects and per-pair margin changes are analyzed in App.~\ref{app:charge}and~\ref{app:quadrant}

\subsection{Ablation Study (RQ3)}
\label{sec:ablation}

\begin{table}[t]
\centering
\small
\setlength{\tabcolsep}{3.0pt}
\renewcommand{\arraystretch}{1.08}
\resizebox{\columnwidth}{!}{%
\begin{tabular}{l cccc >{\hspace{6pt}}c ccc}
\toprule
& \multicolumn{4}{c}{\textbf{Qwen3-8B-Base}} & \multicolumn{4}{c}{\textbf{GLM-4-9B-Base}} \\
\cmidrule(lr){2-5}\cmidrule(lr){6-9}
\textbf{Variant} & F1$\uparrow$ & Rep.$\uparrow$ & G-Ret.$\uparrow$ & PSR$\uparrow$ & F1$\uparrow$ & Rep.$\uparrow$ & G-Ret.$\uparrow$ & PSR$\uparrow$ \\
\midrule
\rowcolor{oursbg} \textbf{PROSE (full)} & 91.48 & 85.25 & 90.76 & 76.34 & 87.17 & 67.37 & 94.02 & 62.16 \\
\quad w/o attenuation~(Eq.~\ref{eq:presume_attenuation}) & 76.44 & 40.84 & 99.68 & 40.52 & 86.82 & 68.15 & 93.05 & 62.36 \\
\quad w/o constraints~(Eq.~\ref{eq:presume_update_objective}) & 57.27 & 70.83 & 41.00 & 21.39 & 35.40 & 99.81 & 2.90 & 2.70 \\
\quad w/o both & 61.16 & 62.72 & 50.57 & 25.12 & 35.78 & 99.81 & 3.67 & 3.47 \\
\bottomrule
\end{tabular}%
}
\caption{Preservation ablation on CAIL-MC. 
}
\label{tab:ablation_preservation}
\end{table}
To investigate which preservation components support selective repair,
we remove target-level attenuation
(Eq.~\ref{eq:presume_attenuation}), parameter-level preservation constraints
(Eq.~\ref{eq:presume_update_objective}), or both, and refit each variant
without retuning the remaining hyperparameters.
Table~\ref{tab:ablation_preservation} shows that removing parameter-level
constraints sharply reduces matched-G retention and PSR on both backbones,
despite substantial repair.
Attenuation has a backbone-dependent effect: removing it substantially
reduces Qwen's repair and PSR while increasing G-Ret., whereas GLM's PSR remains nearly unchanged. Removing both components also yields poor safe repair.
Thus, parameter-level constraints consistently support selectivity, while attenuation improves Qwen's repair--preservation balance.
Three-stage ablations are reported in App.~\ref{app:stage_ablation}.

\subsection{Training and Inference Efficiency (RQ4)}
\label{sec:efficiency}

\begin{figure}[t]
  \centering
  \includegraphics[width=0.9\columnwidth]{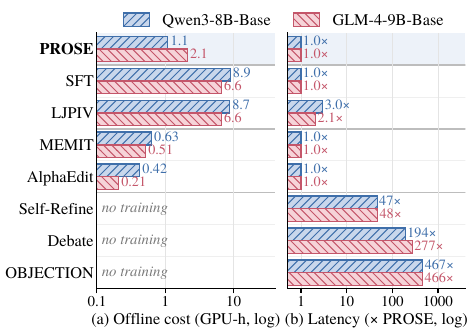}
  \caption{Offline cost on one RTX 4080 SUPER and per-query latency relative to PROSE, both on log scales.}
  \label{fig:cost}
\end{figure}
To investigate the computational cost of correction, we measure offline GPU-hours on one RTX 4080 SUPER and per-query latency relative to \ours.
Fig.~\ref{fig:cost} shows that \ours requires only 1.1/2.1 GPU-hours for Qwen/GLM, reducing offline cost by $3.1$--$8.1\times$ relative to SFT and LJPIV, while retaining ordinary single-pass inference; LJPIV additionally pays $2.1$--$3.0\times$ the latency because it queries the three legal stages separately.
MEMIT and AlphaEdit are cheaper offline but damage the unedited model on both tracks (Sec.~\ref{sec:main_results}), and inference-time reasoning methods need no training but incur $47$--$467\times$ the latency at every query.
Thus, \ours is the only method that combines low cost on both axes with gains over the unedited model in every setting.

\section{Conclusion}
Guilty-bias mitigation requires correcting unsupported verdicts while preserving legally supported ones. We formalize this aim as
preservation-constrained selective repair and introduce \ours, which constrains both activation-target construction and parameter fitting.
We also develop CAIL-MC consisting of two complementary tracks,
evaluating correction at both the matched-pair and evidential-support levels.
Across two backbones, \ours achieves stronger pairwise safe repair than LJP and editing baselines and less supported-verdict damage than a
repair-matched uniform shift. While SFT excels on matched pairs, \ours better responds to changing
evidential support under corpus and construction shift, with lower offline cost and ordinary single-pass inference.

\section*{Limitations}
\label{sec:limitations}
Our study covers charge-conditioned verification based LJP in Chinese criminal
law using two backbones, rather than complete judicial decision-making.
CAIL-MC uses LLM-assisted transformations of adjudicated facts.
Independent legal review reduces, but cannot eliminate, generation
artifacts and uncertainty in legal interpretation. Its coverage-oriented
construction and selection of editable cases do not represent the
distribution of criminal cases in practice.
The Evidence-Sufficiency Generalization Track measures the sufficiency
of factual support available in the supplied text, not evidence
credibility or a judicial standard of proof. Its small source-case
sample and fixed allegation-only template limit the breadth of the
evaluation.
Finally, \ours relies on paired training examples and backbone-specific
configuration choices. Its finite strength search may miss feasible
interventions, and parameter fitting approximates their activation
effects. Preservation is therefore empirical, not guaranteed. 

\section*{Ethical Considerations}
\label{sec:ethics}

CAIL-MC is derived from CAIL2018, LeCaRD, and Criminal-S, whose source texts have undergone privacy protection. Our legal review additionally screened the source and constructed texts for residual identifying
information and newly introduced identifying details. The counterfactual variants are synthetic modifications for benchmark evaluation, not additional claims about individuals in the source cases. Our use of the source data complies with their applicable terms and any release of derived material will follow the corresponding
redistribution conditions.

Our labels concern whether the supplied facts establish a specified charge, not a defendant's factual innocence or overall legal status. The evidence-sufficiency evaluation tests a limited behavioral requirement relevant to the presumption of innocence, rather than compliance with
judicial proof standards. \ours is intended for research, not decisions about criminal responsibility, detention, or sentencing. Improved benchmark performance does not establish legal correctness or fairness,
and any prospective application would require independent validation and qualified legal oversight.

\bibliography{custom}
\appendix
\raggedbottom


\section{CAIL-MC Construction Details}
\label{app:dataset}
CAIL-MC is a benchmark suite with two complementary tracks. The
\textit{Pairwise Repair--Preservation Track} provides matched
charge-established and charge-not-established cases for evaluating
selective repair and preservation. The
\textit{Evidence-Sufficiency Generalization Track} provides controlled
evidence variants from independent corpora for evaluating evidence
sensitivity under corpus and construction shift. The two tracks share
the same principles of fact sanitization, legal review, and outcome
leakage control, but serve different purposes. 

\subsection{Pairwise Repair--Preservation Track}
\label{app:pairwise-track}
We construct this track from
CAIL2018~\citep{xiao2018cail}. We first retain cases with an
identifiable target charge and sufficiently complete facts for
controlled counterfactual editing, removing duplicate, malformed,
or factually incomplete records, which yields 20,055 candidates.
These cases then undergo legal-path planning, counterfactual
generation, automated factual and legal checks, and expert human
verification. Cases for which no legally valid counterpart can be
constructed without altering the core source facts are discarded,
resulting in 11,478 charge-established/charge-not-established
matched pairs spanning 261 Chinese criminal charges. Given a
charge-established case $x_i^{G}$ with charge $c_i$, we construct
$x_i^{NG}$ such that $c_i$ is no longer established while preserving the remaining factual context as much as possible, so that each pair differs only in a legally decisive condition.

\subsubsection{Legal Path Planning}

Because fact descriptions extracted from original judgments may
contain expressions that reveal adjudicative outcomes, we first use DeepSeek-V4-Flash to extract and sanitize the case facts, removing conclusion-revealing expressions. The planning prompt
$P_{\mathrm{PLAN}}$ then identifies candidate counterfactual paths for the cleaned case at three levels: elements of the offense,unlawfulness, and culpability.

Real-world guilty cases are naturally easier to modify by negating an element of the offense. Simply selecting the easiest construction would therefore over-concentrate the dataset on offense-level cases and reduce the coverage of unlawfulness and culpability. We thus first attempt eligible unlawfulness and culpability paths, ranking candidates according to the remaining coverage gap of each legal path, feasibility, and conflict risk. Here, a \textit{legal path} denotes a concrete legal construction mechanism, such as self-defense and necessity under unlawfulness, or mental incapacity and age below criminal responsibility under culpability. \textit{Feasibility} measures whether a complete counterfactual can be naturally constructed from the current case, whereas \textit{conflict risk} measures whether the introduced facts are likely to contradict the original case facts or hard constraints. A case proceeds to offense-level construction only when no reasonable U/C path can satisfy the quality requirements.

\subsubsection{Counterfactual Generation}

Given the selected legal path, the counterfactual generation prompt$P_{\mathrm{GEN}}$ instructs DeepSeek-V4-Flash to construct the corresponding charge-not-established fact description while preserving the original case facts as much as possible and introducing only the information necessary to alter the target legal condition.

Rather than mechanically appending exculpatory descriptions, we require the newly introduced facts to be adapted to the specific case context. For example, a self-defense counterfactual must be constructed around the original actors, course of aggression, and temporal relations, while a culpability-related modification must be consistent with the defendant's specific identity and condition. This design avoids simply inserting fixed exculpatory keywords and makes the generated cases more consistent with realistic legal scenarios. To further prevent shortcut learning, the generated text is prohibited from directly containing conclusion-revealing expressions such as ``not guilty'' or ``does not constitute a crime.''

Each generated sample is then automatically reviewed by DeepSeek-V4-Flash. Samples that fail the review are regenerated according to reviewer feedback for at most two refinement rounds. If a U/C path still fails after refinement, the case is transferred to offense-level construction, where the same generation and review procedure is applied.
\subsubsection{Quality Verification}
To ensure the reliability of the constructed data, we adopt a
two-stage quality-control mechanism combining LLM review with expert
human verification. At the factual level, the automated reviewer
defined by $P_{\mathrm{REV}}$ checks whether the modified case
preserves the main facts and narrative of the original case while
avoiding factual contradictions, excessive modification, and
conclusion leakage. At the legal level, it verifies whether the
designated legal path is correctly instantiated and whether the added
or modified facts are sufficient to make the target charge no longer
established.

After LLM-based construction and automatic verification, we
conducted a full human review of the remaining 12,564 candidate
pairs. The review team consisted of eight junior annotators and
one senior adjudicator. The senior adjudicator holds a doctoral
degree in law, serves as a law professor at a leading university,
and has 13 years of experience in criminal-law research, teaching,
and/or judicial practice. The eight junior annotators were
master's students specializing in criminal law at leading
universities. All had completed systematic coursework in both
the General Part and the Specific Part of Chinese Criminal Law
and had received professional training in criminal-case analysis.

\paragraph{Independent dual review.}The 12,564 candidate pairs produced after LLM-based construction
and automatic verification were randomly divided into four equal
review batches, each containing 3,141 pairs. Each batch was
assigned to two different junior annotators, and each junior
annotator was responsible for one batch. The two annotators
independently reviewed every pair in their assigned batch and
were not shown each other's decisions during the independent
review stage.

For each pair, the reviewers jointly considered the original
charge-established fact, the generated charge-not-established
fact, the target charge, the assigned doctrine, and the designated
legal path. They examined whether the two texts still described
the same underlying case; whether the designated legal path was
correctly instantiated; whether the modified or introduced facts
were legally sufficient to make the target charge no longer
established; whether the counterfactual condition covered all
independently charge-establishing acts in cases involving repeated
or continuous conduct; whether the temporal, behavioral, and
causal relations remained internally consistent; and whether the
counterfactual avoided excessive modification, competing doctrines,
and direct outcome leakage.

Each annotator independently assigned a \textit{retain} or
\textit{discard} decision to every candidate and recorded a brief
reason whenever a candidate was considered invalid.
\paragraph{Expert adjudication.}
When the two junior annotators independently reached the same
decision, that decision was adopted directly. Any candidate on
which the two annotators disagreed was forwarded directly to the
senior criminal-law expert for final adjudication, without an
intermediate discussion stage.

For each disputed candidate, the senior expert re-examined the
target charge, assigned doctrine, designated legal path, original
case, counterfactual case, and the reasons recorded by both junior
annotators, and then made the final decision. A candidate was
discarded when a legally valid charge-not-established counterpart
could not be obtained while preserving the core facts of the
original case. This included cases in which another independently
charge-establishing act remained intact, the counterfactual
contradicted material source facts, the intended legal effect
required introducing an additional doctrine, or the underlying
case had to be substantially rewritten to invalidate the target
charge.


\noindent\textbf{Human-review outcomes.}\quad
Tab.~\ref{tab:human-review-results} summarizes the human-review
outcomes across the four batches. Every one of the 12,564
candidate pairs was independently examined by two annotators with
formal training in criminal law, and every disagreement between
the two reviewers was further adjudicated by the senior expert.
Human verification ultimately discarded 1,086 candidates and
retained 11,478 matched pairs, corresponding to a final retention
rate of 91.36\%.

\begingroup
\setlength{\intextsep}{5pt}
\setlength{\abovecaptionskip}{4pt}
\setlength{\belowcaptionskip}{0pt}

\begin{table}[t]
\centering
\small

\setlength{\tabcolsep}{3.5pt}

\resizebox{\columnwidth}{!}{%
\begin{tabular}{l|c|c|c|c|c}
\toprule
&
\textbf{Total} &
\textbf{Agree} &
\textbf{Disagree} &
\textbf{Discarded} &
\textbf{Retained} \\
\midrule
Batch 1 & 3,141 & 2,779 & 362 & 267 & 2,874 \\
Batch 2 & 3,141 & 2,758 & 383 & 274 & 2,867 \\
Batch 3 & 3,141 & 2,791 & 350 & 261 & 2,880 \\
Batch 4 & 3,141 & 2,744 & 397 & 284 & 2,857 \\
\midrule
\textbf{Total}
& \textbf{12,564}
& \textbf{11,072}
& \textbf{1,492}
& \textbf{1,086}
& \textbf{11,478} \\
\bottomrule
\end{tabular}
}
\caption{
Human verification results across the four review batches.
``Agree'' denotes identical decisions independently made by the
two junior annotators. Every ``Disagree'' case was directly
referred to the senior criminal-law expert for final adjudication.
}
\label{tab:human-review-results}

\end{table}
\endgroup






\subsubsection{Construction Prompts}
 \label{app:cailmc-prompts}


\planboxsimple{
Please sanitize and analyze the following Chinese criminal case.

\textbf{Charge:}\quad
\texttt{\{charge\}}

\textbf{Raw fact:}\quad
\texttt{\{raw\_fact\}}

\textbf{Task 1: Generate \texttt{guilty\_fact}.}

\begin{itemize}
\setlength{\itemsep}{1pt}

\item Preserve the actor, object, time, location, conduct, means,
monetary amount, quantity, result, causal relation, and necessary
mental-state facts.

\item Remove evidentiary listings, procedural history, sentencing
information, adjudicative conclusions, and outcome leakage.

\item Do not omit repeated offenses, continuous offenses, or
multiple acts that would each independently suffice to establish
the current charge.

\end{itemize}

\textbf{Task 2: Extract:}

\begin{itemize}
\setlength{\itemsep}{1pt}

\item \texttt{core\_facts}: the core facts supporting establishment
of the current charge;

\item \texttt{hard\_constraints}: explicit facts that subsequent
counterfactual editing must not directly contradict, such as an
explicitly stated age, an explicit assessment of criminal
responsibility, the offense timeline, the number of repeated
criminal acts, explicit monetary amounts or quantities, an explicit
harmful result, and whether the infringement has already ended.

\end{itemize}

\textbf{Task 3: Evaluate all major counterfactual paths separately.}

\textbf{Important principles:}

\begin{itemize}
\setlength{\itemsep}{1pt}

\item Judge \emph{counterfactual constructibility}, not whether an
exculpatory circumstance actually already exists in the original
case.

\item \texttt{feasibility} evaluates only whether the path can be
constructed through a relatively minimal, case-relevant, and
temporally coherent change.

\item For \texttt{unlawfulness}, any introduced infringement or
danger must occur before or concurrently with the original conduct
and must cover all acts sufficient to establish the current charge.
A later event cannot negate an already completed offense.

\item For \texttt{culpability}, a mistake about a constitutive fact
(e.g., not knowing that an item is stolen property, a drug, a
counterfeit medicine, or a protected object) belongs under
\texttt{offense}, not \texttt{culpability}. A \texttt{culpability}
mistake refers only to an unavoidable mistake of illegality.

\item Independently and fully evaluate \texttt{unlawfulness} and
\texttt{culpability} before evaluating \texttt{offense}. Do not
lower U/C feasibility merely because an \texttt{offense} path is
easier to construct.

\item For every U/C \texttt{legal\_path}, evaluate it from the
perspective that the minimum necessary counterfactual facts may be
introduced. Do not merely check whether the original text already
contains the relevant exculpatory facts.

\item For \texttt{offense}, in principle always search for a single
offense element that can be modified. If the first element is
unsuitable, search for another element rather than assigning
\texttt{offense=0} too readily.

\end{itemize}

\textbf{Feasibility scale:}

\begin{itemize}
\setlength{\itemsep}{1pt}

\item 3 = naturally fits the case with a small change;
\item 2 = can be constructed reasonably;
\item 1 = theoretically constructible but somewhat forced;
\item 0 = clearly unreasonable.

\end{itemize}

\textbf{Output JSON only:}

{\ttfamily\upshape\scriptsize
\{\par
\quad "guilty\_fact": "...",\par
\quad "core\_facts": ["..."],\par
\quad "hard\_constraints": ["..."],\par
\quad "doctrine\_plans": \{\par

\qquad "offense": \{\par
\qquad\quad "candidates": [\par
\qquad\qquad \{\par
\qquad\qquad\quad
"legal\_path": "single\_element\_negation",\par
\qquad\qquad\quad "feasibility": 0,\par
\qquad\qquad\quad "changed\_element": "...",\par
\qquad\qquad\quad "minimal\_change": "...",\par
\qquad\qquad\quad
"full\_exculpation\_possible": true,\par
\qquad\qquad\quad
"conflict\_risk": "low|medium|high",\par
\qquad\qquad\quad "reason": "..."\par
\qquad\qquad \}\par
\qquad\quad ]\par
\qquad \},\par

\qquad "unlawfulness": \{\par
\qquad\quad "candidates": [\par

\qquad\qquad \{\par
\qquad\qquad\quad
"legal\_path": "self\_defense",\par
\qquad\qquad\quad "feasibility": 0,\par
\qquad\qquad\quad "minimal\_change": "...",\par
\qquad\qquad\quad
"full\_exculpation\_possible": true,\par
\qquad\qquad\quad "temporal\_fit": true,\par
\qquad\qquad\quad
"covers\_all\_criminal\_acts": true,\par
\qquad\qquad\quad
"conflict\_risk": "low|medium|high",\par
\qquad\qquad\quad "reason": "..."\par
\qquad\qquad \},\par

\qquad\qquad \{\par
\qquad\qquad\quad
"legal\_path": "necessity",\par
\qquad\qquad\quad "feasibility": 0,\par
\qquad\qquad\quad "minimal\_change": "...",\par
\qquad\qquad\quad
"full\_exculpation\_possible": true,\par
\qquad\qquad\quad "temporal\_fit": true,\par
\qquad\qquad\quad
"covers\_all\_criminal\_acts": true,\par
\qquad\qquad\quad
"conflict\_risk": "low|medium|high",\par
\qquad\qquad\quad "reason": "..."\par
\qquad\qquad \}\par

\qquad\quad ]\par
\qquad \},\par

\qquad "culpability": \{\par
\qquad\quad "candidates": [\par

\qquad\qquad \{\par
\qquad\qquad\quad
"legal\_path": "mental\_incapacity",\par
\qquad\qquad\quad "feasibility": 0,\par
\qquad\qquad\quad "minimal\_change": "...",\par
\qquad\qquad\quad
"full\_exculpation\_possible": true,\par
\qquad\qquad\quad
"conflict\_risk": "low|medium|high",\par
\qquad\qquad\quad "reason": "..."\par
\qquad\qquad \},\par

\qquad\qquad \{\par
\qquad\qquad\quad
"legal\_path": "age\_below\_responsibility",\par
\qquad\qquad\quad "feasibility": 0,\par
\qquad\qquad\quad
"minimal\_change": "...",\par
\qquad\qquad\quad
"full\_exculpation\_possible": true,\par
\qquad\qquad\quad
"conflict\_risk": "low|medium|high",\par
\qquad\qquad\quad "reason": "..."\par
\qquad\qquad \},\par

\qquad\qquad \{\par
\qquad\qquad\quad
"legal\_path":
"unavoidable\_illegality\_mistake",\par
\qquad\qquad\quad "feasibility": 0,\par
\qquad\qquad\quad
"minimal\_change": "...",\par
\qquad\qquad\quad
"full\_exculpation\_possible": true,\par
\qquad\qquad\quad
"conflict\_risk": "low|medium|high",\par
\qquad\qquad\quad "reason": "..."\par
\qquad\qquad \}\par

\qquad\quad ]\par
\qquad \}\par

\quad \}\par
\}\par
}
}


\genboxsimple{
Construct a high-quality guilty--innocent pair according to the
fixed \texttt{doctrine} and fixed \texttt{legal\_path}.

\textbf{Charge:}\quad
\texttt{\{charge\}}

\textbf{Fixed doctrine:}\quad
\texttt{\{doctrine\}}

\textbf{Fixed legal path:}\quad
\texttt{\{legal\_path\}}

\textbf{Suggested minimal change:}\quad
\texttt{\{minimal\_change\}}

\textbf{Planner-specified changed element:}\quad
\texttt{\{planned\_changed\_element\}}

\textbf{Complete guilty fact:}\quad
\texttt{\{guilty\_fact\}}

\textbf{Core facts:}\quad
\texttt{\{core\_facts\}}

\textbf{Hard constraints:}\quad
\texttt{\{hard\_constraints\}}

\textbf{Mandatory requirements:}

\textbf{I. Same-case identity and factual fidelity}

\begin{enumerate}
\setlength{\itemsep}{1pt}

\item \texttt{innocent\_fact} must still describe the same case.

\item Do not introduce unrelated persons, locations, another
criminal event, or a new harmful result.

\item Comply with all \texttt{hard\_constraints}.

\item For repeated criminal conduct, the counterfactual must apply
to every act sufficient to establish the current charge; it must
not explain away only one occurrence.

\item Output a complete and coherent factual paragraph.

\end{enumerate}

\textbf{II. \texttt{offense}}

\begin{itemize}
\setlength{\itemsep}{1pt}

\item Change only one element of the offense.

\item You may locally modify facts directly related to that element.

\item If \texttt{guilty\_fact} contains a sentence that directly
conflicts with the new element, locally replace that sentence rather
than retaining both A and not-A.

\item Apart from the designated \texttt{changed\_element}, preserve
all other material facts as much as possible.

\item Do not use self-defense, necessity, mental disorder, age, or
any other doctrine.

\end{itemize}

\textbf{III. \texttt{unlawfulness}}

\texttt{self\_defense}:

\begin{itemize}
\setlength{\itemsep}{1pt}

\item State the ongoing unlawful infringement, the protected legal
interest, and that the original conduct was undertaken to stop that
infringement; directly connect the timing and purpose.

\item The infringement must be ongoing at the time of the original
conduct. Do not use a later event to justify earlier completed
criminal conduct.

\end{itemize}

\texttt{necessity}:

\begin{itemize}
\setlength{\itemsep}{1pt}

\item State the specific ongoing danger, the protected legal
interest, the absence of an obviously less harmful alternative, and
the direct connection between the original conduct and avoidance of
the danger.

\item The danger must be present at the time of the original conduct,
and the justificatory circumstance must cover all acts sufficient
to establish the current charge.

\end{itemize}

\textbf{IV. \texttt{culpability}}

\texttt{mental\_incapacity}:

\begin{itemize}
\setlength{\itemsep}{1pt}

\item State that a forensic psychiatric evaluation was conducted.

\item Make explicit that, at the time of the conduct, the person was
unable to recognize or control the conduct.

\item The condition must match the timing of all relevant criminal
acts.

\end{itemize}

\texttt{age\_below\_responsibility}:

\begin{itemize}
\setlength{\itemsep}{1pt}

\item State the person's specific age at the time of the conduct.

\item Do not directly conflict with an explicitly stated adulthood
fact in \texttt{guilty\_fact}.

\item The age must be consistent with the age-of-criminal-
responsibility rule applicable to the current charge.

\end{itemize}

\texttt{unavoidable\_illegality\_mistake}:

\begin{itemize}
\setlength{\itemsep}{1pt}

\item The actor knows the relevant constitutive facts.

\item The mistake concerns whether the conduct is legally prohibited.

\item Provide a concrete, credible, and unavoidable reason for the
mistake, such as an explicit but erroneous response from a competent
authority.

\item Do not place mistakes about constitutive facts---such as not
knowing that an item is stolen property, a drug, a counterfeit
medicine, or a protected object---under this path.

\end{itemize}

\textbf{V. No conclusion leakage}

Do not include conclusion statements such as ``not guilty,''
``does not constitute a crime,'' ``legally bears no criminal
responsibility,'' ``the court finds,'' or ``criminal responsibility
shall be pursued.''

\textbf{Output JSON only:}

{\ttfamily\upshape\scriptsize
\{\par
\quad "doctrine": "\{doctrine\}",\par
\quad "legal\_path": "\{legal\_path\}",\par
\quad "innocent\_fact": "...",\par
\quad "changed\_element": "...",\par
\quad "added\_elements": ["..."],\par
\quad "preserved\_elements": ["..."],\par
\quad "template\_family": "..."\par
\}\par
}
}


\reviewboxsimple{
You are a quality reviewer for a Chinese criminal-law
counterfactual dataset. Conduct a substantive review. Do not merely
match keywords, and do not automatically reject a candidate simply
because a newly introduced counterfactual fact was not stated in
the original text.

\textbf{Charge:}\quad
\texttt{\{charge\}}

\textbf{Fixed doctrine:}\quad
\texttt{\{doctrine\}}

\textbf{Fixed legal path:}\quad
\texttt{\{legal\_path\}}

\textbf{Guilty fact:}\quad
\texttt{\{guilty\_fact\}}

\textbf{Core facts:}\quad
\texttt{\{core\_facts\}}

\textbf{Hard constraints:}\quad
\texttt{\{hard\_constraints\}}

\textbf{Candidate innocent fact:}\quad
\texttt{\{innocent\_fact\}}

\textbf{Check the following dimensions:}

\begin{enumerate}
\setlength{\itemsep}{1pt}

\item \texttt{preservation}: apart from the fixed counterfactual
change, are the material facts substantially preserved?

\item \texttt{path\_consistency}: does the candidate actually
instantiate the designated \texttt{legal\_path}?

\item \texttt{doctrine\_sufficiency}: are the added facts sufficient
to satisfy the designated doctrine?

\item \texttt{full\_exculpation}: does the counterfactual cover
every act sufficient to establish the charge, especially in
repeated or continuous offending?

\item \texttt{temporal\_consistency}: do the justificatory or
exculpatory facts occur before or concurrently with the criminal
conduct, rather than using a later event to explain away an already
completed offense?

\item \texttt{counterfactual\_conflict}: does the candidate avoid
simultaneously retaining contradictory A and not-A facts?

\item \texttt{hard\_constraint\_consistency}: does the candidate
comply with all explicit \texttt{hard\_constraints}?

\item \texttt{plausibility}: are the added facts directly related
to the actors, conduct, and setting of the case, and do they
constitute a relatively minimal change rather than a complex new
story?

\item \texttt{single\_doctrine}: does the candidate avoid mixing
in another doctrine?

\item \texttt{no\_conclusion\_leakage}: is the candidate free of
conclusion leakage?

\item \texttt{non\_mechanical}: is the counterfactual integrated
naturally rather than mechanically appended as a template?

\end{enumerate}

For \texttt{offense}, local replacement of the single
\texttt{changed\_element} is allowed. Therefore, do not mark
\texttt{preservation} as failed solely because a local fact about
mental state, object nature, or a similar element has changed.

\textbf{Output JSON only:}

{\ttfamily\upshape\scriptsize
\{\par
\quad
"decision": "pass|rewrite\_required|switch\_path",\par
\quad
"preservation": "pass|fail",\par
\quad
"path\_consistency": "pass|fail",\par
\quad
"doctrine\_sufficiency": "pass|fail",\par
\quad
"full\_exculpation": "pass|fail",\par
\quad
"temporal\_consistency": "pass|fail",\par
\quad
"counterfactual\_conflict": "pass|fail",\par
\quad
"hard\_constraint\_consistency": "pass|fail",\par
\quad
"plausibility": "pass|fail",\par
\quad
"single\_doctrine": "pass|fail",\par
\quad
"no\_conclusion\_leakage": "pass|fail",\par
\quad
"non\_mechanical": "pass|fail",\par
\quad
"defects": ["..."],\par
\quad
"rewrite\_instruction": "..."\par
\}\par
}
}

\subsection{Evidence-Sufficiency Generalization Track}
\label{app:evidence-track}
This track is constructed from LeCaRD~\citep{ma2021lecard} and Criminal-S
\citep{hu2018few}. It is an independent evaluation component of CAIL-MC,
designed to test whether a model follows the sufficiency of the presented
evidence under both corpus shift and construction shift. 

For each selected source case and candidate charge, we sanitize the fact
description, remove outcome-revealing language, and identify the facts
that support the charge. We then construct variants that preserve the
underlying allegation and case context while changing the amount of
charge-supporting evidence. Each variant is assigned to one of four
conditions:

\begin{itemize}
\item \textbf{Full}: the available facts contain the complete
charge-supporting evidence and establish the candidate charge;

\item \textbf{Partial-sufficient}: only part of the supporting evidence
is retained, but the remaining evidence is still sufficient to establish
the charge;

\item \textbf{Partial-insufficient}: some relevant facts remain, but at
least one indispensable condition is unavailable, so the candidate
charge is not established under the benchmark annotation;

\item \textbf{Allegation-only}: the charge or alleged conduct is retained
without supporting evidence needed to establish the charge.
\end{itemize}

The first two conditions therefore have label $1$ for the specified
charge, while the latter two have label $0$. These labels concern only
whether the specified charge is established from the presented facts; they do not constitute a complete determination of factual innocence or the defendant's overall legal status.

The construction follows the same quality principles as the Pairwise Repair--Preservation Track. GPT-5.6 Sol first extracts the source facts and charge-relevant conditions, after which a generation step retains or removes evidence according to the assigned condition without introducing unrelated events or explicit verdict cues. Automated review checks narrative consistency, preservation of the core allegation, compliance with the intended evidence condition, and absence of label leakage.

The source case and its variants are then reviewed jointly by two criminal-law-trained annotators. They verify the intended evidence condition, the corresponding charge-established or charge-not-established label, and consistency with the original case. Disagreements are
adjudicated by a senior legal reviewer. Variants that cannot satisfy the assigned evidence condition without changing the underlying case are discarded. Human-review outcomes are shown in Tab.~\ref{tab:gradient-human-review}.

\begin{table}[t]
\centering
\small
\caption{
Human verification results for the evidence-sufficiency gradient.
``Agree'' denotes identical retain/discard decisions independently
made by the two criminal-law graduate reviewers.
Every ``Disagree'' case is directly referred to the senior
criminal-law expert for final adjudication.
}
\setlength{\tabcolsep}{4pt}
\resizebox{\columnwidth}{!}{%
\begin{tabular}{l|c|c|c|c|c}
\toprule
\textbf{Source} &
\textbf{Total} &
\textbf{Agree} &
\textbf{Disagree} &
\textbf{Discarded} &
\textbf{Retained} \\
\midrule
LeCaRD
& 80
& 65
& 15
& 10
& 70 \\

Criminal-S
& 200
& 194
& 6
& 0
& 200 \\
\midrule
\textbf{Total}
& \textbf{280}
& \textbf{259}
& \textbf{21}
& \textbf{10}
& \textbf{270} \\
\bottomrule
\end{tabular}%
}
\label{tab:gradient-human-review}
\end{table}

%

\definecolor{gradgreen}{HTML}{4F7865}
\definecolor{gradbggreen}{HTML}{EFF6F2}

\NewDocumentCommand{\gradboxsimple}
{O{Evidence-ablation construction prompt $P_{\mathrm{GRAD}}$}+m}{%
    \promptboxbase{gradgreen}{gradbggreen}{#1}{#2}%
}

%
\newcommand{\LeCaRDAgree}{\textit{TBD}}
\newcommand{\LeCaRDDisagree}{\textit{TBD}}
\newcommand{\CriminalAgree}{\textit{TBD}}
\newcommand{\CriminalDisagree}{\textit{TBD}}

\newcommand{\GradientAgreeTotal}{\textit{TBD}}
\newcommand{\GradientDisagreeTotal}{\textit{TBD}}

\subsubsection{Evidence-Ablation Construction Prompt}
\label{app:gradient-prompt}

\gradboxsimple{

\textbf{Task.}
You are constructing a guilty-side evidence-ablation gradient for
a legal judgment experiment.

\medskip
\textbf{Target charge:}
\emph{\{charge\}}

\medskip
\textbf{FULL:}
\emph{\{full\}}

\medskip
FULL is the original \texttt{guilty\_fact} and is fixed.
You must never rewrite FULL.
You only construct
\textbf{Partial-sufficient},
\textbf{Partial-insufficient}, and
\textbf{Zero}.

\medskip
\textbf{Partial-sufficient}

Target label: \emph{charge established}.

Remove only background, repetitive, redundant, or non-essential
information from FULL.
After deletion, the remaining facts must still sufficiently
support the target charge.
Do not remove all factual support required for any necessary
condition.
A necessary condition does not need to appear verbatim: if it can
still be reasonably inferred from other retained facts, it remains
supported.

Do not add, supplement, or fabricate any fact.
Preserve the original sentences, order, fluency, and writing style
of FULL as much as possible.
If deletion creates an ungrammatical sentence, only minimal
coherence-preserving rewriting is allowed.
Do not add new legal conclusions such as ``the crime is
established'' or ``guilty.''

For this level, additionally record:
(1) the exact spans removed from FULL and why they can be removed;
(2) the necessary supports that remain and their retained factual
bases; and
(3) why the resulting evidence remains sufficient.

\medskip
\textbf{Partial-insufficient}

Target label:
\emph{not established by the provided factual material}.

Delete factual support required for at least one necessary
condition of the target charge.
After deletion, the other retained facts must not substitute for,
independently support, or reasonably imply the same necessary
condition.
Do not mechanically treat a condition as unsupported merely
because it is not stated verbatim; consider all retained facts and
whether that condition can still be reasonably inferred.

Insufficiency must be created by deleting facts.
Only minimal rewriting necessary for grammatical coherence is
allowed.
Do not add exculpatory, negating, contradictory, or
innocence-supporting facts.
Do not transform the deletion of supporting evidence into an
affirmative statement that the deleted fact did not occur.
Do not add new legal conclusions such as ``does not constitute the
crime'' or ``not guilty.''

For this level, record:
(1) \texttt{missing\_necessary\_support}, identifying the missing
necessary support;
(2) \texttt{deleted\_or\_rewritten\_spans}, identifying the exact
spans of FULL that were deleted or minimally rewritten;
(3) \texttt{why\_remaining\_cannot\_support}, explaining why the
remaining facts cannot continue to support the missing condition;
and
(4)
\texttt{alternative\_retained\_support\_sufficient}, which must be
\texttt{false}.

\medskip
\textbf{Zero}

Target label:
\emph{not established by the provided factual material}.

Do not retain any evidentiary support from FULL.
Do not retain any case-specific fact concerning conduct, time,
location, amount, object, consequence, or mental state.
Retain only a neutral accusation naming the target charge.

The output text must exactly follow the fixed template:

\texttt{The defendant is accused of being involved in \{charge\}}

\medskip
\textbf{Control of the two partial conditions}

Subject to the sufficient/insufficient distinction above,
keep the deletion amounts as similar as possible;
keep the resulting text lengths as similar as possible;
keep linguistic fluency as similar as possible; and
preserve the narrative order and writing style of FULL as much as
possible.
Legal evidentiary sufficiency takes priority over mechanically
matching text length.

\medskip
\textbf{Information restriction}

Use only the target charge and FULL supplied in this prompt.

Do not use, infer, or introduce information from
\texttt{innocent\_fact}, \texttt{doctrine},
\texttt{legal\_path}, \texttt{changed\_element},
\texttt{added\_elements}, or \texttt{preserved\_elements} from
the original matched pair.
The doctrine used to construct the original pair does not imply
that factual support removed from the guilty side should follow
the same doctrine.

\medskip
\textbf{Output format}

Return exactly one JSON object, with no Markdown and no additional
explanation.

The object must contain
\texttt{partial\_sufficient},
\texttt{partial\_insufficient}, and
\texttt{zero}.

\texttt{partial\_sufficient} must contain
\texttt{text},
\texttt{removed\_spans},
\texttt{retained\_necessary\_support}, and
\texttt{why\_still\_sufficient}.
Each removed span must report the exact
\texttt{original\_span} from FULL and a brief
\texttt{reason}.

\texttt{partial\_insufficient} must contain
\texttt{text},
\texttt{missing\_necessary\_support},
\texttt{deleted\_or\_rewritten\_spans},
\texttt{why\_remaining\_cannot\_support}, and
\texttt{alternative\_retained\_support\_sufficient}.
Each deleted or minimally rewritten span must specify its
\texttt{operation}, exact
\texttt{original\_span}, optional
\texttt{replacement}, and
\texttt{reason}.
When the operation is \texttt{delete},
\texttt{replacement} must be empty.

\texttt{zero} contains only the fixed neutral accusation text.
}

\section{Diagnosis Details}
\label{app:diagnosis}


\subsection{Diagnostic Protocol}
\label{app:diag}

All diagnostic interventions use the main verification prompt and verdict readout, with the backbone in bf16, the verdict logits read in fp32, and inputs truncated to 2,048 tokens; model weights stay frozen. Source pairs are training pairs whose two members $\theta_0$ classifies correctly, and targets are the charge-not-established members of strict false-guilty development pairs (153 / 185 / 286 Offense / Unlawfulness / Culpability pairs with Qwen).

\paragraph{Layer-wise transfer.}
For each legal stage and each of the 36 layers, the contrasts of Eq.~\ref{eq:diag_pair_direction} are averaged over the source pairs of that stage and normalized to a unit direction $v_d^\ell$. In a single forward pass, $\alpha v_d^\ell$ with $\alpha=3$ is added to the block output at the three answer-adjacent positions $\mathcal T$ of a target. \emph{Forward} adds $v_d^\ell${, \emph{Reverse} adds $-v_d^\ell$ to the same target}, and \emph{Norm-matched random} adds three independent Gaussian directions per target, each rescaled to the norm of the forward displacement. NG-margin reduction is the clean margin minus the intervened margin. Confidence intervals are case-cluster bootstrap intervals (2,000 resamples), with the random repeats clustered by case.

\paragraph{Feasibility and damage.}
At Layer~22, a case-specific direction is reconstructed in the stage-specific or shared affine geometry (Eq.~\ref{eq:presume_affine}), and a case is feasible if a strength within the bounded search moves its NG margin below the buffer $\tau$. Paired-G damage applies the same target-reaching displacement to the matched charge-established member.

\subsection{Full Preservation Projection}
\label{app:diag_full}

\begin{table}[t]
\centering
\small
\setlength{\tabcolsep}{4pt}
\resizebox{\columnwidth}{!}{%
\begin{tabular}{llcc}
\toprule
Stage & Geometry & Without & With full projection \\
\midrule
\multirow{2}{*}{Offense} & Stage-specific & 70.6 (108) & 17.6 (27) \\
 & Shared & 70.6 (108) & 20.3 (31) \\
\midrule
\multirow{2}{*}{Unlawfulness} & Stage-specific & 83.2 (154) & 15.7 (29) \\
 & Shared & 76.8 (142) & 19.5 (36) \\
\midrule
\multirow{2}{*}{Culpability} & Stage-specific & 75.9 (217) & 10.8 (31) \\
 & Shared & 76.9 (220) & 10.1 (29) \\
\bottomrule
\end{tabular}}
\caption{Repair-target feasibility (\%, feasible pairs in parentheses) on Qwen3-8B-Base without and with full projection onto the complement of the supported-guilty gradient subspace ($\rho=0$ in Eq.~\ref{eq:presume_attenuation}).}
\label{tab:diag_full}
\end{table}

Full projection removes the entire component of the repair direction that lies in the supported-guilty gradient subspace $P_d^\ell$. Tab.~\ref{tab:diag_full} shows that it lowers feasibility from 70.6--83.2\% to 10.1--20.3\%: the removed component carries much of the repair signal. The damage comparison in Fig.~\ref{fig:diag_qwen}c is therefore restricted to the pairs feasible under both settings, whose counts are those in parentheses. This trade-off motivates the soft attenuation of PROSE ($\rho>0$), which keeps part of the overlapping component.

\subsection{Cross-Stage Transfer of Hidden-state Contrasts}
\label{app:diag_cross}
\begin{figure}[t]
  \centering
  \includegraphics[width=0.82\columnwidth]{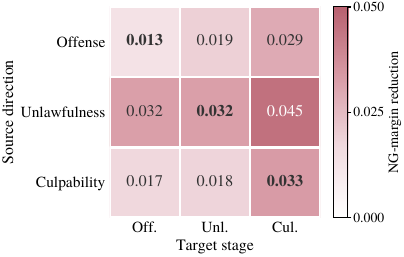}
  \caption{Cross-stage transfer at Layer~22 on Qwen3-8B-Base. Each cell adds the contrast direction of a source stage to cases of a target stage and reports the mean NG-margin reduction; the diagonal (bold) gives each direction's effect on its own stage.}
  \label{fig:causal_l22}
\end{figure}

To test whether the stage-specific contrasts share structure, we add the unit contrast direction of each source stage, scaled by {$\alpha=0.5$}, at Layer~22 to the {held-out cases} of each target stage and measure the mean NG-margin reduction (Fig.~\ref{fig:causal_l22}). All nine effects are positive, so every stage direction moves cases of every stage toward charge not established, consistent with the shared geometry that development selection chooses for Offense and Unlawfulness. The Unlawfulness direction transfers most strongly (0.032--0.045), matching the stage ablation, where the Unlawfulness update alone reaches 63.58 PSR on Culpability pairs (Tab.~\ref{tab:stage_all}). The Culpability direction is the most stage-selective (0.033 on its own stage vs.\ 0.017--0.018 elsewhere), consistent with its stage-specific geometry, while Offense has the weakest own-stage effect (0.013), in line with Offense being the hardest stage to repair in isolation.

\noindent
\begin{minipage}{\columnwidth}
    \centering
    \includegraphics[
        width=0.80\linewidth
    ]{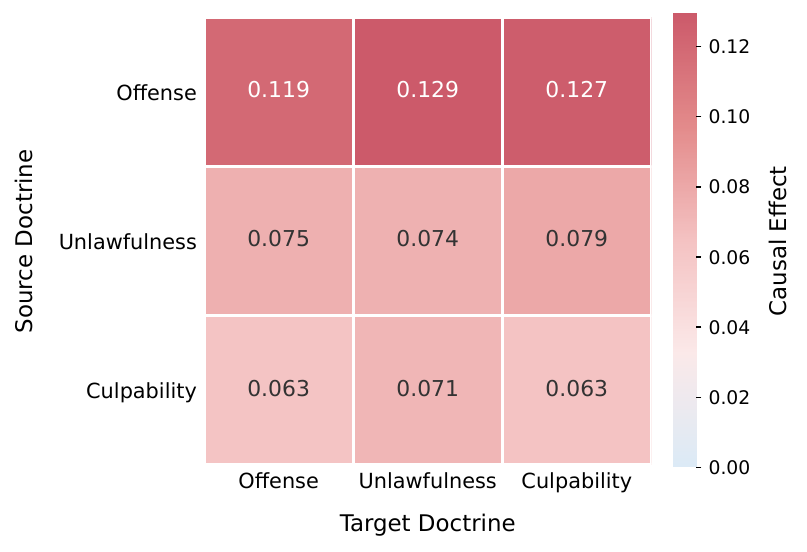}

    \captionof{figure}{
    \textbf{Cross-stage transfer on GLM-4-9B-Base.}
    Each cell reports the mean NG-margin reduction when the
    hidden-state contrast learned from the source doctrine (row)
    is applied to held-out cases from the target doctrine (column)
    at Layer~20. Positive values indicate movement toward charge
    not established.
    }
    \label{fig:causal_glm_l20}
\end{minipage}
\par

Fig.~\ref{fig:causal_glm_l20} again shows positive transfer in
all nine source--target combinations, confirming that the learned
contrast directions contain a substantial cross-stage component.
On GLM-4-9B-Base, the Offense direction is strongest across all
three target stages (0.119--0.129), followed by Unlawfulness
(0.074--0.079) and Culpability (0.063--0.071). Moreover, several
cross-stage effects are comparable to or larger than the
corresponding within-stage effects. Thus, both backbones exhibit
shared cross-stage contrast structure, while the degree and form
of stage specialization remain backbone dependent.

\subsection{Transfer of Hidden-state Contrasts}
\label{app:diag_glm}
\label{app:contrast-transfer}


\noindent
\begin{minipage}{\columnwidth}
    \centering
    \includegraphics[
        width=\linewidth
    ]{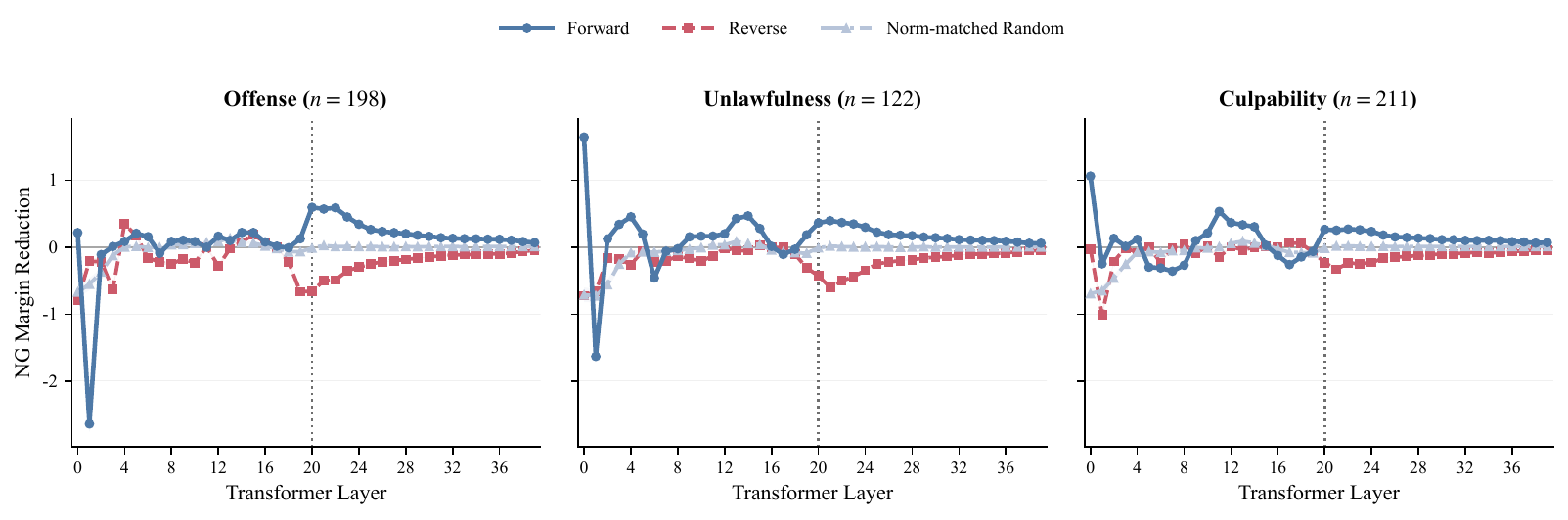}

    \captionof{figure}{
    \textbf{Layer-wise hidden-state contrast transfer on
    GLM-4-9B-Base.}
    Each panel reports NG-margin reduction on held-out strict
    false-guilty pairs for one legal stage.
    Forward applies the learned contrast direction, Reverse its
    negation, and Norm-matched Random a random direction with the
    same norm. Positive values indicate movement toward charge not
    established; the dotted line marks Layer~20.
    }
    \label{fig:glm-layer-transfer}
\end{minipage}

\par\vspace{4pt}

Fig.~\ref{fig:glm-layer-transfer} shows a qualitatively similar
directional-transfer pattern to that observed on Qwen3-8B, despite
a different layer profile. Around the middle layers, the Forward
direction produces positive NG-margin reduction across all three
legal stages, whereas Reverse tends to produce the opposite effect
and the norm-matched random control is substantially weaker.

 \subsection{Repair Feasibility and Preservation}
 \label{app:repair-feasibility}

\begin{figure}[t]
    \centering
    \includegraphics[
        width=\columnwidth
    ]{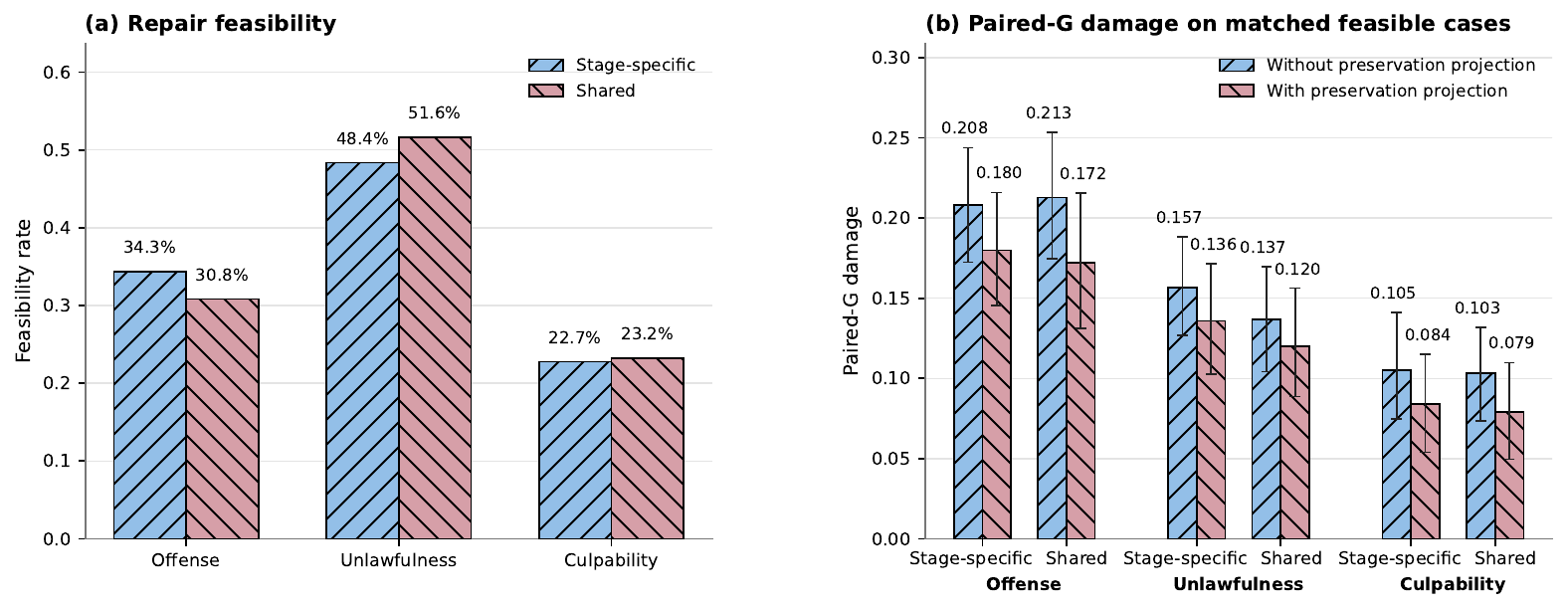}
    \caption{
    \textbf{Repair feasibility and preservation on GLM-4-9B-Base.}
    \textbf{(a)} Repair feasibility under stage-specific and shared
    contrast geometry for the three legal stages.
    \textbf{(b)} Paired-G margin damage on matched feasible cases,
    before and after preservation projection.
    Lower paired-G damage indicates better preservation of the
    supported charge-established counterpart.
    }
    \label{fig:glm-repair-preservation}
\end{figure}

As shown in Fig.~\ref{fig:glm-repair-preservation}(a), GLM-4-9B-Base
has lower overall repair feasibility than Qwen3-8B, ranging from
22.7\% to 51.6\% across legal stages and geometry choices.
Nevertheless, the transferred contrast geometry reaches the target
for a non-trivial fraction of false-guilty cases. 

In
Fig.~\ref{fig:glm-repair-preservation}(b), preservation projection
reduces paired-G damage in all six stage/geometry comparisons,
with absolute reductions of 0.017--0.041. Thus, the GLM results
reproduce the same qualitative repair--preservation tension:
repair-effective directions are transferable, but explicit
preservation is needed to limit damage to the matched supported
decision.

\section{Details of \ours}
\label{app:method}
This appendix expands the target construction and parameter update in
Sec.~\ref{sec:method}. We retain the following notation: $\theta_0$ fixes eligibility
and preservation labels, whereas $\vartheta$ denotes the current model during
successive stage updates. All contrasts, gradients, and module-input keys
used for an update are extracted under $\vartheta$.

\subsection{Contrast Geometry and Preservation Attenuation}
\label{app:method_geometry}

\paragraph{Affine contrast reconstruction.}
At stage $d$ and layer $\ell$, let $N_d$ be the number of training pairs
resolved correctly by $\theta_0$, and let $h$ be the hidden width. Their
contrasts $q_j^\ell\in\mathbb R^h$ are recomputed under $\vartheta$ using
Eq.~\ref{eq:diag_pair_direction}. Define the centered contrast matrix
\begin{equation}
\small
\begin{aligned}
\bar q_d^\ell&=\frac{1}{N_d}\sum_j q_j^\ell,\\
Q_d^\ell&=\big[q_j^\ell-\bar q_d^\ell\big]_j
\in\mathbb R^{h\times N_d}.
\end{aligned}
\label{eq:app_centered_contrasts}
\end{equation}
The sum and columns range over these correctly resolved stage-$d$ pairs.
The leading $r_q$ left singular vectors of $Q_d^\ell$ form $U_d^\ell$;
the requested rank is capped by the available SVD dimensions. The affine
reconstruction in Eq.~\ref{eq:presume_affine} is the least-squares
projection of the target contrast onto
$\bar q_d^\ell+\operatorname{span}(U_d^\ell)$:
\begin{equation}
\small
\widehat q_i^\ell
=\underset{a\in\bar q_d^\ell+\operatorname{span}(U_d^\ell)}
{\operatorname{argmin}}\ \|q_i^\ell-a\|_2^2.
\label{eq:app_affine_projection}
\end{equation}
Writing $a=\bar q_d^\ell+U_d^\ell w$, orthonormality gives
$w^\star=(U_d^\ell)^\top(q_i^\ell-\bar q_d^\ell)$, recovering the
reconstruction in the main text.
When the centered matrix is numerically zero, the reconstruction reduces to
the mean. A candidate with fewer than two source pairs is not fitted. A
pooled geometry, when selected, uses correctly resolved training pairs from
all stages in the same construction; target eligibility remains stage-specific.

\paragraph{Intervention and gradient aggregation.}
The displacement $\delta$ is added to the block output at every position in
$\mathcal T$, the three answer-adjacent prompt positions defined in diagnosis.
These are tokenizer positions in the Chinese answer cue, rather than tokens
of its English translation. For a shared displacement, the derivative is
\begin{equation}
\small
\left.\nabla_\delta\widetilde s_{\vartheta}
(x,c;\delta,\ell)\right|_{\delta=0}
=\sum_{t\in\mathcal T}
\nabla_{h_{\vartheta,t}^{\ell}(x,c)}s_{\vartheta}(x,c).
\label{eq:app_shared_gradient}
\end{equation}
Thus, contrasts average hidden-state differences across positions, whereas
the gradient sums their sensitivities to the shared displacement. Gradients
are computed with the model weights frozen. For preservation, we stack the
unnormalized gradients of base-correct charge-established training cases at
stage $d$ and retain the leading $r_p$ left singular vectors as $P_d^\ell$;
this gradient matrix is not centered.

For a nonzero target gradient, $g_i^\ell$ minimizes the first-order margin
change over unit displacements:
\begin{equation}
\small
\begin{aligned}
\min_{\|b\|_2=1}
&\left(\left.\nabla_\delta\widetilde s_i^{\mathrm{NG}}
\right|_{\delta=0}\right)^\top b\\
&=-\left\|\left.\nabla_\delta\widetilde s_i^{\mathrm{NG}}
\right|_{\delta=0}\right\|_2.
\end{aligned}
\label{eq:app_local_descent}
\end{equation}
Here, $\widetilde s_i^{\mathrm{NG}}$ abbreviates the intervened margin of
target $i$ under $\vartheta$. This explains the local descent component;
mixing it with a contrast and attenuating it need not retain this optimality.

\paragraph{Soft attenuation.}
For $\rho>0$, the unnormalized attenuation in
Eq.~\ref{eq:presume_attenuation} admits the following interpretation:
\begin{equation}
\small
a_i^\ell=\underset{a}{\operatorname{argmin}}
\left\{\|a-v_i^\ell(\lambda)\|_2^2
+\rho^{-1}\|(P_d^\ell)^\top a\|_2^2\right\}.
\label{eq:app_attenuation_objective}
\end{equation}
Setting its derivative to zero and using
$(P_d^\ell)^\top P_d^\ell=I$ gives
\begin{equation}
\small
\begin{aligned}
a_i^\ell&=\left[I+\rho^{-1}P_d^\ell(P_d^\ell)^\top\right]^{-1}
v_i^\ell(\lambda)\\
&=\left[I-\frac{P_d^\ell(P_d^\ell)^\top}{1+\rho}\right]
v_i^\ell(\lambda).
\end{aligned}
\label{eq:app_attenuation_solution}
\end{equation}
We then set $u_i^\ell=\operatorname{unit}(a_i^\ell)$. The limit $\rho=0$
is full orthogonal removal; increasing $\rho$ weakens attenuation. This
interpretation applies before normalization and does not guarantee verdict
retention.

\subsection{Bounded Strength Search}
\label{app:method_search}

For an eligible target, write
$f_i(\beta)=\widetilde s_{\vartheta}
(x_i^{\mathrm{NG}},c_i;\beta u_i^\ell,\ell)$.
The cap $B_i$ is defined in Eq.~\ref{eq:presume_beta_cap} using the
reconstruction norm before direction mixing and attenuation. This keeps the
admissible range comparable across preservation settings.

\paragraph{Endpoint test and refinement.}
For a valid direction and positive cap, we first evaluate $f_i(B_i)$.
The search proceeds only if this value is finite and at most $\tau$.
Otherwise, it reports no detected target. Starting from
$L_{i,0}=0$ and $H_{i,0}=B_i$, we perform $J$ bisection steps. At step $j$,
let $M_{i,j}=(L_{i,j}+H_{i,j})/2$ and update
\begin{equation}
\small
(L_{i,j+1},H_{i,j+1})=
\begin{cases}
(L_{i,j},M_{i,j}),&f_i(M_{i,j})\leq\tau,\\
(M_{i,j},H_{i,j}),&\text{otherwise}.
\end{cases}
\label{eq:app_bisection}
\end{equation}
A nonfinite midpoint takes the second branch. We re-evaluate the final upper
endpoint and retain $\beta_i^\star=H_{i,J}$ only when its margin is finite
and satisfies the target. The displacement is then
$t_i=\beta_i^\star u_i^\ell$ as in Eq.~\ref{eq:presume_beta}.
The implementation uses $J=8$ refinement steps.

Each accepted refinement halves the interval width, leaving
$H_{i,J}-L_{i,J}=B_i/2^J$. If the response has a single monotone crossing
with a nonfeasible lower endpoint, the returned endpoint approximates that
crossing from above. Neural margins need not be monotone: an infeasible
capped endpoint can conceal a feasible interior region, and bisection does
not establish a global minimum. Consequently, feasibility here means that
this finite search detects a target-reaching displacement.

\paragraph{Numerical safeguards.}
Normalization uses a numerical floor of $10^{-12}$. Before searching, we
discard nonfinite or near-zero attenuated directions and directions whose
retained norm ratio
$\|a_i^\ell\|_2/\|v_i^\ell(\lambda)\|_2$ falls below a configured
threshold (default $0.25$). When the contrast--gradient mixture nearly
cancels, the gradient-assisted implementation falls back to the negative
margin-gradient direction. Invalid directions and failed searches remain
in the candidate denominator used to report feasibility. Only detected
targets with positive strength contribute key--target columns to fitting;
a stage with no such targets receives no update.

The batched search checks zero strength using the stored reference margin,
not a fresh margin under $\vartheta$. For strict false-guilty targets this
reference margin is positive, so the zero-strength branch does not remove
cases repaired by an earlier stage. Capped and midpoint evaluations use the
current model. This implementation should therefore not be interpreted as
explicitly skipping every previously repaired target.

\subsection{Module Keys and Closed-Form Fitting}
\label{app:method_closed_form}

\paragraph{Joint projection parameterization.}
For attention output and MLP down projections with input widths
$m_{\mathrm{attn}}$ and $m_{\mathrm{mlp}}$, respectively, the block notation
in the main text is horizontal concatenation:
\begin{equation}
\small
\begin{aligned}
W&=[W_{\mathrm{attn}},W_{\mathrm{mlp}}],\\
k&=[k_{\mathrm{attn}}^\top,k_{\mathrm{mlp}}^\top]^\top,\\
\Delta Wk&=\Delta W_{\mathrm{attn}}k_{\mathrm{attn}}
+\Delta W_{\mathrm{mlp}}k_{\mathrm{mlp}}.
\end{aligned}
\label{eq:app_joint_parameterization}
\end{equation}
Here, $W\in\mathbb R^{h\times m}$ with
$m=m_{\mathrm{attn}}+m_{\mathrm{mlp}}$. A single-module update uses the
corresponding block alone. Each intervened position contributes one column
of $K_{\mathcal F}\in\mathbb R^{m\times n_{\mathcal F}}$ and repeats
its pair's $t_i$ in $T_{\mathcal F}\in\mathbb R^{h\times n_{\mathcal F}}$.
Preservation matrices $K_a$ use the same positions and module order, so
$n_{\mathcal F}$ and $n_a$ count columns rather than cases.

Keys are frozen during fitting. The sum in
Eq.~\ref{eq:app_joint_parameterization} is a local fitting surrogate:
updating attention also changes the subsequent MLP input. Accordingly,
least-squares target fit need not equal the hidden-state or verdict change
under a full post-update forward pass.

\paragraph{Weighted ridge solution.}
To express Eq.~\ref{eq:presume_update_objective} as a single ridge
problem, define repair blocks
$\mathsf A_{\mathcal F}=K_{\mathcal F}/\sqrt{n_{\mathcal F}}$ and
$\mathsf Y_{\mathcal F}=T_{\mathcal F}/\sqrt{n_{\mathcal F}}$.
For each nonempty preservation pool $a\in\{\mathrm G,\mathrm O,\mathrm X\}$,
define $\mathsf A_a=\sqrt{\omega_a/n_a}\,K_a$ and a matching zero target
block $\mathsf Y_a=0$. Concatenate these blocks into $\mathsf A$ and
$\mathsf Y$, omitting empty pools. If $N$ is their total column count,
$\mathsf A\in\mathbb R^{m\times N}$ and
$\mathsf Y\in\mathbb R^{h\times N}$. Then
\begin{equation}
\small
\mathcal L(\Delta W)=
\|\Delta W\mathsf A-\mathsf Y\|_F^2+\mu\|\Delta W\|_F^2.
\label{eq:app_ridge_objective}
\end{equation}
Pool normalization prevents larger pools from receiving greater weight
solely because they contain more columns. Differentiating gives
\begin{equation}
\small
\begin{aligned}
\tfrac12\nabla_{\Delta W}\mathcal L
&=(\Delta W\mathsf A-\mathsf Y)\mathsf A^\top+\mu\Delta W,\\
\Delta W_{\mathrm{LS}}(\mathsf A\mathsf A^\top+\mu I_m)
&=\mathsf Y\mathsf A^\top.
\end{aligned}
\label{eq:app_normal_equation}
\end{equation}
For fixed activation targets and preservation keys, $\mu>0$ makes the
ridge objective strictly convex, yielding a unique closed-form solution:
\begin{equation}
\small
\begin{aligned}
\Delta W_{\mathrm{LS}}
&=\mathsf Y\mathsf A^\top
(\mathsf A\mathsf A^\top+\mu I_m)^{-1}\\
&=\mathsf Y(\mathsf A^\top\mathsf A+\mu I_N)^{-1}\mathsf A^\top.
\end{aligned}
\label{eq:app_ridge_solution}
\end{equation}
Here, $I_m$ and $I_N$ are identity matrices of the indicated dimensions.
The implementation uses the second, dual form with a Cholesky solve rather
than explicitly computing an inverse. It retains factors
$L=\mathsf Y(\mathsf A^\top\mathsf A+\mu I_N)^{-1}$ and
$R=\mathsf A^\top$, so $\Delta W_{\mathrm{LS}}=LR$ need not be materialized.

\subsection{Low-Rank Realization and Update Budget}
\label{app:method_realization}
We retain a numerical approximation to the leading singular components of
$LR$. The factorized SVD computation uses a fixed Gaussian sketch determined
by the run seed, with oversampling of $16$ and one power iteration.
Writing the resulting factors as $U_W$, $S_W$, and $V_W$, the rank-$r_W$ update is
\begin{equation}
\small
\Delta W_{r_W}
=U_{W,:r_W}S_{W,:r_W}V_{W,:r_W}^\top.
\label{eq:app_rank_truncation}
\end{equation}
The retained rank is capped by the available dimensions. This operation
approximates a truncation of the ridge solution; it is not an exact solution
to a rank-constrained version of Eq.~\ref{eq:presume_update_objective}.
The captured singular-value energy is measured relative to
$\|LR\|_F^2=\operatorname{tr}[(L^\top L)(RR^\top)]$.

Let $b$ index the selected projection modules, and let
$\Delta W_{r_W,b}$ be the corresponding column block. The budget is
enforced separately for each module through a common scaling factor:
\begin{equation}
\small
\begin{aligned}
c_W&=\max_b\frac{\|\Delta W_{r_W,b}\|_F}{\|W_b\|_F},\\
\eta_W&=\min\{1,\varepsilon/c_W\},\\
W_b'&=W_b+\eta_W\Delta W_{r_W,b}.
\end{aligned}
\label{eq:app_module_budget}
\end{equation}
For a numerically zero update we set $\eta_W=1$; weight norms use the same
numerical floor as normalization. Every module then satisfies the relative
Frobenius-norm budget $\varepsilon$, and updates already below the budget
are not amplified. The applied block update in the main text is
$\Delta W=\eta_W\Delta W_{r_W}$.

Fitting quality can be checked through
$\|\Delta W K_{\mathcal F}-T_{\mathcal F}\|_F/\|T_{\mathcal F}\|_F$
and preservation drift through
$\|\Delta W K_a\|_F/\sqrt{n_a}$. These are fixed-key diagnostics.
Actual verdict repair and retention are measured with the updated model,
because truncation, rescaling, and downstream nonlinearities can change the
realized effect.

Before application, the implementation checks that the fitting pool is
nonempty, the retained singular-value energy meets its configured minimum,
and each module satisfies both the selected budget and an emergency norm
ceiling, allowing a numerical tolerance of $10^{-6}$. A failed check skips
the update. These are numerical acceptance checks, not tests of behavioral
preservation.

\subsection{Stage Execution and Data Separation}
\label{app:method_execution}

\paragraph{Fixed membership and preservation pools.}
The reference predictions define
$\mathcal E_{\mathrm{sFG}}^{\mathrm{train}}(\theta_0)$, the correctly resolved
source pairs, and preservation membership once. For a stage-$d$ update,
$\mathrm G$ contains base-correct charge-established training cases at that
stage; $\mathrm O$ contains charge-not-established members of correctly
resolved stage-$d$ pairs. Across other stages, $\mathrm X$ includes
base-correct charge-established members and the charge-not-established
members of correctly resolved pairs. Size-limited training pools are selected
once under the fixed run configuration and held fixed during fitting.
The gradient bank and the weight-fitting protection pool may have different
sizes, but both satisfy these reference-defined eligibility criteria.
The eligible $\mathrm G$ pool includes target counterparts; a capped fitting
pool need not contain every counterpart.
\paragraph{Successive offline updates.}
Each selected stage configuration specifies its layer, source geometry,
direction mixture, preservation strength, search limits, fitting penalties,
rank, and update budget. The reported configurations apply updates in the
order \textsc{Offense}, \textsc{Unlawfulness}, then \textsc{Culpability}.
Starting with $\vartheta=\theta_0$, each update recomputes source and target
contrasts, gradients, and keys under the current $\vartheta$, constructs
search-detected targets, solves the ridge problem, and applies the bounded
low-rank update. Cached features are cleared after a weight change. Reference
membership remains fixed even when the current model's predictions change.
The final current model is $\theta'$. The offline phase can therefore contain
several stage updates, while deployment requires only one ordinary forward
pass and no matched counterpart.

\paragraph{Selection and evaluation.}
Contrast statistics, gradients, targets, and fitting keys use training data.
Development data select configurations; candidate parameter changes are
restored before an independent trial. The selected configurations are fixed
before final fitting and evaluation. Neither test track supplies fitting
examples or selection signals. In particular, the Evidence-Sufficiency
Generalization Track is evaluation-only. Numerical defaults reported above
describe solver controls; they should not be read as the selected
backbone-specific hyperparameters.
\section{Experimental Setup Details}
\label{app:setup}

\subsection{Pair-Level Metric Definitions}
\label{app:pair_metrics}

Pair-level metrics are reported only for the Pairwise Repair--Preservation
Track. For a split $S$ and backbone, the strict false-guilty cohort
$\mathcal E_{\mathrm{sFG}}^S(\theta_0)$ is selected once using the reference
model $\theta_0$ and then kept fixed for all evaluated models. Let
$N_S = |\mathcal E_{\mathrm{sFG}}^S(\theta_0)|$. The evaluated model's
prediction is denoted by $\hat y_{\theta}(x,c)$, where $1$ indicates that the
specified charge is established and $0$ that it is not established.

We define NG Repair as
\begin{equation}
\small
\mathrm{Rep.}(\theta;S)
=
\frac{1}{N_S}
\sum_{i\in\mathcal E_{\mathrm{sFG}}^S(\theta_0)}
\mathbf{1}\!\left[
\hat y_{\theta}(x_i^{\mathrm{NG}},c_i)=0
\right].
\label{eq:pair_rep}
\end{equation}

Matched-G retention is
\begin{equation}
\small
\mathrm{G\mbox{-}Ret.}(\theta;S)
=
\frac{1}{N_S}
\sum_{i\in\mathcal E_{\mathrm{sFG}}^S(\theta_0)}
\mathbf{1}\!\left[
\hat y_{\theta}(x_i^{\mathrm{G}},c_i)=1
\right].
\label{eq:pair_gret}
\end{equation}

Pairwise safe repair requires both the unsupported member to be corrected and
the supported counterpart to remain charge-established:
\begin{equation}
\small
\begin{aligned}
\mathrm{PSR}(\theta;S)
=
\frac{1}{N_S}
\sum_{i\in\mathcal E_{\mathrm{sFG}}^S(\theta_0)}
\mathbf{1}\![
\hat y_{\theta}(x_i^{\mathrm{NG}},c_i)&=0\\
\;\land\;
\hat y_{\theta}(x_i^{\mathrm{G}},c_i)&=1].
\end{aligned}
\label{eq:pair_psr}
\end{equation}

Rep.\ measures the marginal repair ability, whereas G-Ret.\ measures
preservation of the matched charge-established decision. PSR is their
intersection and therefore evaluates selective repair directly.
Consequently, a high Rep.\ with low G-Ret.\ indicates that repair is obtained
by damaging supported decisions, while a high G-Ret.\ with low Rep.\ indicates
that the correction is too conservative. PSR rewards only pairs satisfying
both requirements. All three metrics use the same fixed cohort and denominator;
pairs that remain unrepaired or cannot produce a feasible target are not
removed from evaluation.

These metrics measure within-pair repair and preservation. They do not by
themselves establish preservation of unrelated cases or evidence sensitivity;
those properties are evaluated separately with case-level and
condition-level results.


\subsection{Evaluation Uncertainty}

\paragraph{Confidence intervals.}
The reported intervals are $\pm1.96$ bootstrap standard errors over 1,000 resamples (seed 42), i.e., normal-approximation 95\% intervals. The resampling unit is a matched pair for the Pairwise Repair--Preservation Track and a source case, with all four evidence variants retained together, for the Evidence-Sufficiency Generalization Track. Pair-level metrics are computed on the fixed reference-defined strict false-guilty cohort, restricted to the resampled pairs that belong to it. Models are not refitted during resampling, so these intervals quantify evaluation-sample uncertainty, not variation across fitting seeds.

\paragraph{Paired comparisons.}
For comparisons with the unedited model, both models are evaluated on the same case-level bootstrap resamples, and we report as a one-sided $p$-value the fraction of resamples in which PROSE does not exceed the unedited model \citep{koehn2004statistical}. Individual-method intervals are not used as a test of between-method differences.

\subsection{Baselines}
\label{app:baselines}

\paragraph{LJPIV.}
LJPIV~\citep{zhang2025beyond} extends LJP to non-guilty outcomes and judges the offense elements, unlawfulness, and culpability in sequence. We fine-tune it on the same CAIL-MC training pairs, separately for each backbone, with its original hyperparameters: LoRA ($r=16$, $\alpha=32$, dropout 0.05) on all seven projections of every layer, learning rate $5\times10^{-5}$, batch size 16, a cosine schedule, and 3 epochs. At inference, it queries the three stages with its stage-wise prompt and reads each stage verdict from the logits.

\paragraph{Inference-time methods.}
We follow the implementations of \citet{objection}, with prompts adapted to Chinese charge-conditioned verification.
\begin{itemize}
\item \emph{Self-Refine}~\citep{madaan2023selfrefine} revises an initial verdict with feedback from a neutral critic (3 calls per case).
\item \emph{Debate-Feedback}~\citep{chen2025debate} aggregates three debaters over one revision round (8 calls).
\item \emph{OBJECTION}~\citep{objection} inserts an adversarial lawyer agent into each legal stage and stops once a stage is negated. Facts are truncated to 2,048 tokens within a context of 8,192, and the generation limits are 256 tokens for schema extraction, 192 for the judge, 320 for the lawyer, and 128 for the final answer.
\end{itemize}

\paragraph{Model editors.}
MEMIT~\citep{meng2023mass} edits factual associations by treating MLP projections as key--value memories: it optimizes a target hidden value for each edit request and writes all requests into the weights at once with a closed-form least-squares update, regularized by the second moment of keys from preserved knowledge. AlphaEdit~\citep{fang2024alphaedit} builds on this formulation but projects the update onto the null space of the preserved keys, so that the outputs for preserved inputs are left unchanged by construction. We adapt both to binary verdict repair. Their defining components are kept, and everything task-specific is matched to PROSE.
\begin{itemize}
\item \emph{Requests, prompt, and readout.} The edit requests are all strict false-guilty training pairs of each stage, without subsampling, scored with PROSE's prompt and readout.
\item \emph{Location.} Both editors write to the MLP down-projection of PROSE's edited layer, their native write location, at the same three prompt positions; the stages are edited in the order Offense, Unlawfulness, Culpability.
\item \emph{Target values.} As in the original methods, each request's target is an additive vector on the down-projection output, optimized by back-propagation (Adam, learning rate 0.1, 25 steps, weight decay $10^{-3}$) with its norm clamped to 4 times the unedited output norm. The object likelihood is replaced by $\mathrm{ReLU}(s-\tau)^2$ with PROSE's endpoint $\tau$.
\item \emph{MEMIT preservation.} The key second moment is estimated from {128} generic training references, which contain both members of their pairs and exclude the edit requests. The covariance weight is $3\times10^{3}$ and the ridge {$10^{-2}$}; the update is solved exactly in Woodbury form.
\item \emph{AlphaEdit preservation.} The null space of the key second moment of all non-request training cases (both members) is computed once on the unedited model, with eigenvalue threshold $10^{-2}$ and ridge $10^{-2}$. Keys of earlier stage edits are accumulated as protected history for later stages, as in its sequential setting.
\end{itemize}
Editor hyperparameters are tuned on the development split, and the test sets are loaded only after all edits are fitted. The editors differ from PROSE in how targets are obtained and in what is protected: generic reference keys rather than pools built around the matched pairs; they use neither PROSE's rank cap nor its update budget.

\paragraph{SFT.}
SFT follows the Fine-Tuning-Direct variant of LJPIV, which predicts the final verdict without trichotomous reasoning, and receives the same adapter capacity as our LJPIV baseline. Given PROSE's verification prompt, the model is trained to output the verdict token of both members of all 9,181 training pairs (18,362 cases, 263 tokens on average), with a full-vocabulary cross-entropy loss at the answer position only. LoRA ($r=16$, $\alpha=32$, dropout 0.05, no bias) is applied to all seven projections (query, key, value, output, gate, up, and down) of all 36 layers of Qwen, giving 43,646,976 trainable parameters (0.53\% of the model). Following LJPIV, we train with AdamW without weight decay, learning rate $5\times10^{-5}$, a cosine schedule without warmup, effective batch size 16 (micro-batch 1 with 16 accumulation steps), gradient clipping at 1.0, and 3 epochs (3,444 steps), keeping the backbone in bf16 and the LoRA parameters in fp32 with gradient checkpointing. Before training, we verified that the training-time margins match the evaluation scorer (maximum absolute difference below 0.1); the adapter is merged before evaluation.

\paragraph{Uniform-shift control.}
The control of Sec.~\ref{sec:selectivity} leaves the model unchanged and subtracts one constant $\xi$ from the verdict margin of every test case. $\xi$ is set so that the shift repairs the same fraction of the base-misclassified charge-not-established CAIL-MC test cases as PROSE (85.3\% with Qwen and 67.7\% with GLM), giving $\xi=0.69$ for Qwen and $\xi=0.17$ for GLM. Supported-guilty damage is the fraction of base-correct charge-established test cases whose verdict flips.

\subsection{Implementation of PROSE}
\label{app:hyper}

\paragraph{Configuration.}
PROSE edits the attention-output and MLP down-projections of Layer~22 (Qwen) and Layer~20 (GLM) at the last three prompt positions. The solver, low-rank realization, and update budget are described in Appendix~\ref{app:method}. 

\paragraph{Stage execution.}
On both backbones, the three stage updates are fit successively in the order Offense, Unlawfulness, Culpability. Each update is fit on the model edited by the preceding stages, with contrasts, gradients, and keys recomputed after every weight change, while eligibility and preservation membership stay fixed under the unedited model (Appendix~\ref{app:method_execution}).

\paragraph{Subsampling.}
Geometry and preservation examples are drawn deterministically from the training split with the run seed: 64 cases for the supported-guilty gradient subspace $P^{\ell}_d$, 96 for $\mathcal{P}_G$, 96 for $\mathcal{P}_X$, and 64 for $\mathcal{P}_O$.

\paragraph{Runs.}
PROSE involves no gradient-based training: given the training split and the selected configuration, each stage update is obtained by a closed-form ridge solve followed by rank truncation and rescaling. Following MEMIT and AlphaEdit, which are likewise closed-form editors, we report a single fitting run (seed~42).

\paragraph{Model selection.}
For each backbone and legal stage, the candidates combine an edited layer in the middle-layer region identified by the diagnosis, a contrast geometry (stage-specific or shared across stages), and values from the search space of Tab.~\ref{tab:hyper}, including the contrast--gradient mixture $\lambda$. A candidate is admissible if, on the development split, it fits targets for its stage, retains at least 87\% of the base-correct charge-established, same-stage, and cross-stage predictions, keeps at least 90\% of the fitted update after rank truncation, and stays within its update budget. Admissible candidates are ranked lexicographically by PSR, Rep., mean absolute margin drift on the preservation examples, and binary KL on the same examples, with ties broken toward smaller $\kappa$, $\varepsilon$, and $r_W$. All hyperparameters are selected by this procedure on the development split, and the test sets play no role in selection.

\begin{table}[!t]
\centering
\small
\setlength{\tabcolsep}{4pt}
\resizebox{\columnwidth}{!}{%
\begin{tabular}{lll}
\toprule
Symbol & Description & Value / range \\
\midrule
$\ell$ & Edited layer & 20, 22 \\
-- & Edited modules & attn-out + MLP down \\
-- & Contrast geometry & stage-specific or shared \\
$\tau$ & Target margin endpoint & $[-0.02, -0.005]$ \\
$\lambda$ & Contrast--gradient mixture & $[0, 1]$ \\
$r_q$ / $r_p$ & Contrast / preservation rank & 128 / 64 \\
$\rho$ & Attenuation strength & 4 \\
$\beta_{\max}$ & Global strength cap & 128 \\
$\kappa$ & Case-relative strength cap & $[8, 24]$ \\
$\omega_G$ & Supported-guilty pool weight & $[0.5, 4]$ \\
$\omega_X$ & Cross-stage pool weight & $[2.5, 5]$ \\
$\omega_O$ & Same-stage pool weight & $[5, 10]$ \\
$\mu$ & Ridge regularization & 0.1 \\
$r_W$ & Update rank & $[64, 128]$ \\
$\varepsilon$ & Relative update budget & $[0.15, 0.55]$ \\
\bottomrule
\end{tabular}}
\caption{Hyperparameters of PROSE across both backbones. Ranges give the development-split search space; single values are fixed. Configurations are selected per backbone and legal stage on the development split.}
\label{tab:hyper}
\end{table}

\section{Evidence Sensitivity on CAIL-MC}
\label{app:evidence}

\begin{figure}[t]
  \centering
  \includegraphics[width=\columnwidth]{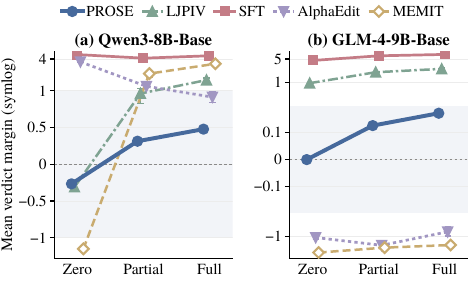}
  \caption{Mean verdict margin on the 1,150 charge-established CAIL-MC test facts rewritten at three evidence levels, with (a) Qwen3-8B-Base and (b) GLM-4-9B-Base. The vertical axis is linear within $[-1,1]$ in (a) and $[-0.2,0.2]$ in (b) (shaded) and logarithmic beyond; error bars are 95\% confidence intervals.}
  \label{fig:evidence_cailmc}
\end{figure}

\paragraph{Construction.}
To test evidence sensitivity without corpus shift, we rewrite the 1,150 charge-established CAIL-MC test facts at three evidence levels, following the evidence-ablation construction and annotation procedure of the Evidence-Sufficiency Generalization Track (App.~\ref{app:evidence-track}): \emph{Zero} states only that the defendant is accused of the charge, \emph{Partial} keeps part of the objective facts, and \emph{Full} keeps all of them. All texts are scored with the main protocol. Fig.~\ref{fig:evidence_cailmc} plots the mean margins of the same five methods as Fig.~\ref{fig:evidence}; Tables~\ref{tab:evidence_rates} and~\ref{tab:evidence} give the rates and the exact margins.

\paragraph{The pattern of Track~2 holds in-domain.}
PROSE is again the only method that, on both backbones, keeps the accusation-only margin at or below zero and the full-fact margin above it, rising from $-$0.27 to 0.48 with Qwen and from $-$0.0001 (95\% CI $[-0.004, 0.004]$) to 0.17 with GLM. SFT stays far above the boundary at every level: with Qwen its margin is highest for the accusation alone (4.84 vs.\ 4.59 for the full facts), and with GLM it remains above 4.5 even for the accusation alone. With Qwen, AlphaEdit also peaks at Zero, and LJPIV and MEMIT separate Zero from Full; with GLM, however, LJPIV places every level above zero and both editors place every level below it.

\begin{table}[!t]
\centering
\small
\setlength{\tabcolsep}{4pt}
\begin{tabular}{lcccc}
\toprule
& \multicolumn{3}{c}{Judged established (\%)} & \\
\cmidrule(lr){2-4}
Method & Zero$\downarrow$ & Partial & Full$\uparrow$ & $\Delta s\uparrow$ \\
\midrule
Unedited & 93.8 & 94.8 & 97.7 & 0.58 \\
Uniform shift & 0.2 & 34.8 & 69.4 & 0.58 \\
SFT & 100.0 & 98.3 & 99.0 & $-$0.25 \\
\rowcolor{oursbg} PROSE & 10.3 & 76.0 & 86.1 & 0.74 \\
\bottomrule
\end{tabular}
\caption{Share of the rewritten CAIL-MC facts judged as establishing the charge with Qwen3-8B-Base, and the mean margin change from Zero to Full ($\Delta s$). The uniform shift is the matched-repair control of Sec.~\ref{sec:selectivity}.}
\label{tab:evidence_rates}
\end{table}

\begin{table}[!t]
\centering
\small
\setlength{\tabcolsep}{2.5pt}
\resizebox{\columnwidth}{!}{%
\begin{tabular}{llccc}
\toprule
& Level & Unedited & SFT & PROSE \\
\midrule
\multirow{3}{*}{Qwen} & Zero & 0.304 [0.293, 0.315] & 4.835 [4.755, 4.921] & $-$0.266 [$-$0.281, $-$0.251] \\
& Partial & 0.568 [0.546, 0.590] & 4.130 [4.003, 4.252] & 0.314 [0.290, 0.339] \\
& Full & 0.888 [0.863, 0.911] & 4.589 [4.468, 4.703] & 0.478 [0.451, 0.504] \\
\midrule
\multirow{3}{*}{GLM} & Zero & 0.059 [0.055, 0.063] & 4.531 [4.389, 4.674] & $-$0.000 [$-$0.004, 0.004] \\
& Partial & 0.179 [0.168, 0.191] & 6.251 [5.944, 6.564] & 0.126 [0.117, 0.135] \\
& Full & 0.272 [0.260, 0.284] & 6.878 [6.557, 7.205] & 0.172 [0.163, 0.181] \\
\bottomrule
\end{tabular}}
\caption{Mean verdict margin [95\% bootstrap CI] of the rewritten CAIL-MC facts at each evidence level.}
\label{tab:evidence}
\end{table}

\paragraph{PROSE keeps the response to evidence.}
Given the accusation alone, unedited Qwen establishes the charge for 93.8\% of the texts, nearly as often as with the full facts (97.7\%). PROSE lowers this rate to 10.3\% while still establishing 86.1\% of the fully evidenced texts, and it steepens the margin increase from Zero to Full from 0.58 to 0.74. The uniform shift reaches a lower Zero rate (0.2\%) only by also rejecting 30.6\% of the fully evidenced texts, and by construction it leaves the margin increase unchanged.

\paragraph{No threshold repairs SFT.}
SFT establishes the charge for nearly every text at every level, and its mean margin is lower for Partial (4.13) and Full (4.59) than for Zero (4.84), so its margin does not grow with evidence ($\Delta s=-0.25$). A different threshold cannot fix this: we raise the SFT threshold until its Zero-level rate is closest to that of PROSE, which happens at $t=6.25$ given tied margins (12.5\% vs.\ 10.3\%). The Partial and Full rates then drop to 12.7\% and 18.7\%, so fully evidenced texts are rejected almost as often as accusation-only ones, and on the CAIL-MC test pairs FNR rises from 0.26 to 52.61 (Acc 73.65, PSR 37.60). SFT's verdicts therefore do not encode evidence sufficiency at any operating point.

\section{Three-Stage Update Ablation}
\label{app:stage_ablation}

\begin{table}[!t]
\centering
\small
\setlength{\tabcolsep}{2.4pt}
\resizebox{\columnwidth}{!}{%
\begin{tabular}{lcccccc|cccc}
\toprule
& Acc & F1 & FGR & FNR & Rep. & G-Ret. & Off. & Unl. & Cul. & All \\
\midrule
\multicolumn{11}{l}{\emph{Qwen3-8B-Base}}\\
O & 72.30 & 70.11 & 54.78 & 0.61 & 7.46 & 100.00 & 26.02 & 1.10 & 3.83 & 7.46 \\
U & 87.96 & 87.87 & 20.35 & 3.74 & 63.53 & 96.76 & 31.71 & 74.03 & 63.58 & 60.29 \\
C & 85.91 & 85.69 & 26.43 & 1.74 & 51.70 & 99.19 & 31.71 & 17.13 & 77.96 & 50.89 \\
OU & 86.26 & 86.11 & 24.00 & 3.48 & 59.00 & 97.08 & 48.78 & 67.96 & 52.40 & 56.24 \\
OC & 82.39 & 81.94 & 33.48 & 1.74 & 40.52 & 99.03 & 44.72 & 16.57 & 50.80 & 39.55 \\
UC & 91.87 & 91.87 & 6.61 & 9.65 & 88.17 & 89.47 & 51.22 & 82.32 & 85.94 & 77.96 \\
\rowcolor{oursbg} OUC & 91.48 & 91.48 & 8.43 & 8.61 & 85.25 & 90.76 & 56.91 & 77.35 & 83.39 & 76.34 \\
\midrule
\multicolumn{11}{l}{\emph{GLM-4-9B-Base}}\\
O & 78.35 & 77.91 & 35.65 & 7.65 & 23.94 & 97.10 & 29.44 & 15.32 & 21.03 & 22.59 \\
U & 76.87 & 76.06 & 41.57 & 4.70 & 14.67 & 99.61 & 6.67 & 29.84 & 12.15 & 14.48 \\
C & 82.48 & 82.21 & 29.74 & 5.30 & 36.87 & 98.65 & 17.22 & 8.06 & 68.22 & 36.10 \\
OU & 81.43 & 81.23 & 29.04 & 8.09 & 37.64 & 96.53 & 35.56 & 42.74 & 31.31 & 35.52 \\
OC & 85.00 & 84.96 & 20.09 & 9.91 & 55.98 & 94.98 & 39.44 & 20.97 & 80.84 & 52.12 \\
UC & 84.74 & 84.60 & 24.78 & 5.74 & 47.10 & 98.07 & 17.22 & 34.68 & 76.64 & 45.95 \\
\rowcolor{oursbg} OUC & 87.17 & 87.17 & 15.04 & 10.61 & 67.37 & 94.02 & 42.22 & 53.23 & 84.11 & 62.16 \\
\bottomrule
\end{tabular}}
\caption{Stage ablation on CAIL-MC (\%). O / U / C denote the Offense, Unlawfulness, and Culpability updates; OUC is PROSE. The right block gives PSR on the strict false-guilty test pairs of each stage and on all of them.}
\label{tab:stage_all}
\end{table}

\begin{figure}[t]
  \centering
  \includegraphics[width=\columnwidth]{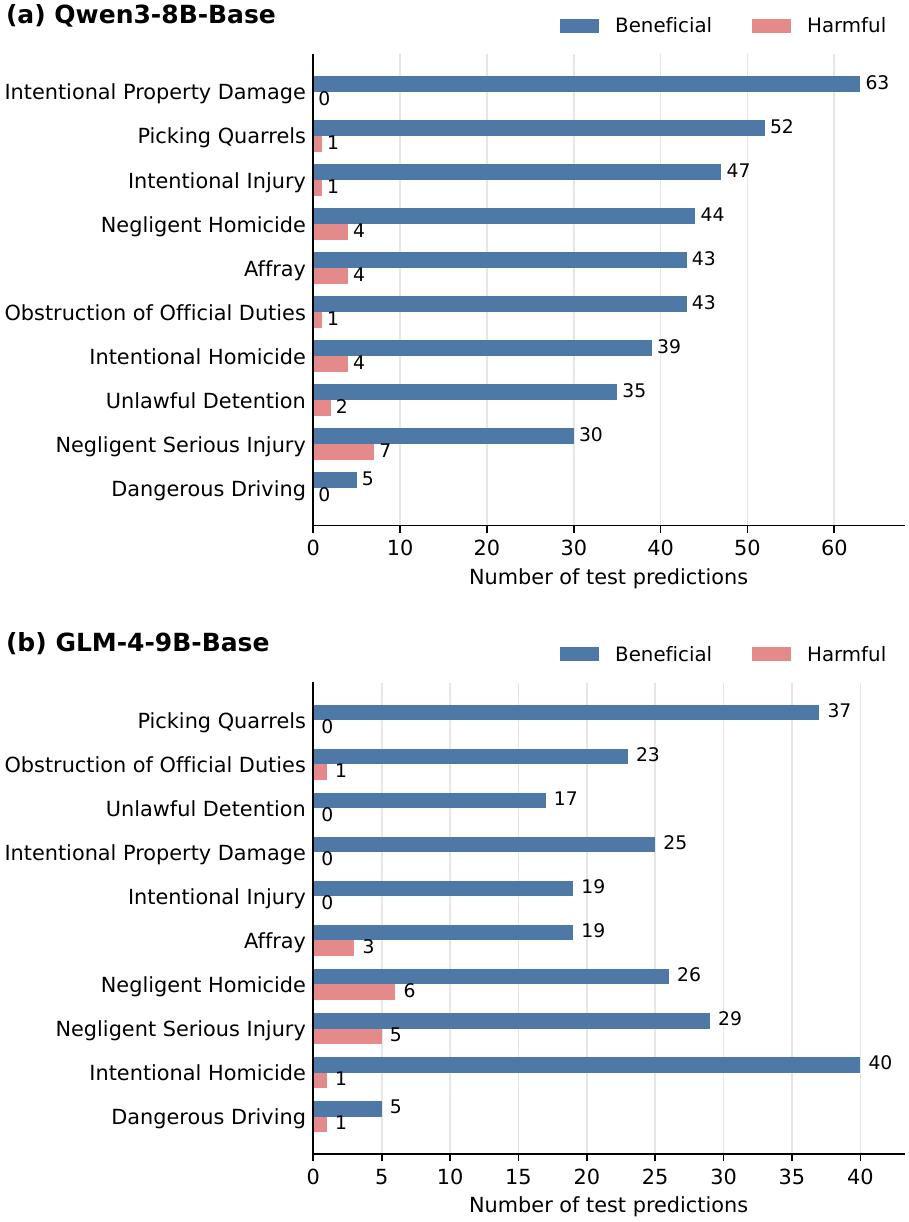}
  \caption{Effect of editing on the ten most frequent charges in the CAIL-MC test split. \emph{Beneficial}: test predictions that editing flips from incorrect to correct; \emph{harmful}: predictions it flips from correct to incorrect.}
  \label{fig:charge}
\end{figure}

\begin{figure}[t]
  \centering
  \includegraphics[width=\columnwidth]{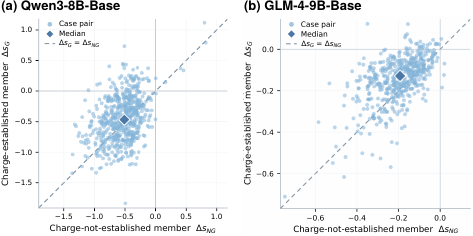}
  \caption{Change in verdict margin of the charge-not-established member ($\Delta s_{NG}$) and of its charge-established counterpart ($\Delta s_{G}$) for each strict false-guilty CAIL-MC test pair under PROSE, on (a) Qwen3-8B-Base and (b) GLM-4-9B-Base. The diamond marks the medians; a uniform output shift would place every pair at a single point on the dashed diagonal.}
  \label{fig:shift_scatter}
\end{figure}
\begin{figure}[t]
  \centering
  \includegraphics[width=\columnwidth]{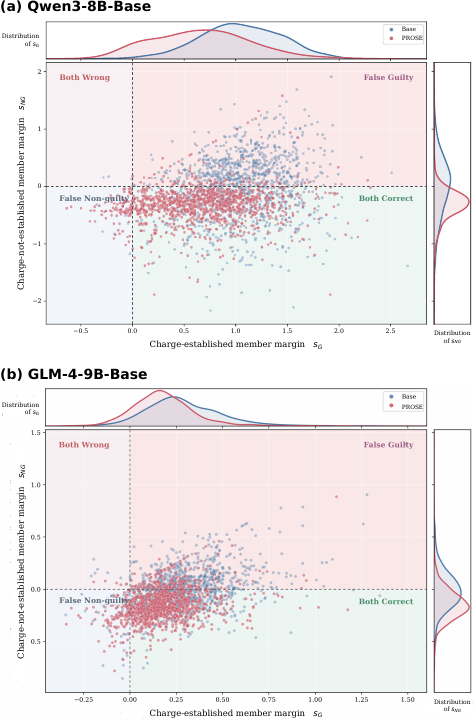}
  \caption{Verdict margins of the charge-established member ($s_{G}$, horizontal) and the charge-not-established member ($s_{NG}$, vertical) for every CAIL-MC test pair under the base model and PROSE, on (a) Qwen3-8B-Base and (b) GLM-4-9B-Base. The dashed lines are the decision boundaries, so the four quadrants are the four joint pair states; the marginal panels show the distribution of each margin.}
  \label{fig:quadrant}
\end{figure}
To examine whether the three legal stages provide complementary repair coverage, we retain only the Offense (O), Unlawfulness (U), or Culpability
(C) update, their pairwise combinations (OU, OC, and UC), or all three updates (OUC). All other parts of \ours remain unchanged. Tab.~\ref{tab:stage_all}
reports case-level metrics and pair-level repair and retention, with PSR also broken down by the changed legal stage and reported over all strict
false-guilty pairs.

The effects are stage- and backbone-dependent rather than monotonic. Combining stages generally broadens coverage, and OUC gives the strongest pooled PSR on GLM. On Qwen, UC has a slightly higher pooled PSR than OUC, whereas OUC performs better on offense-stage pairs. Thus, no fixed subset
dominates across backbones and legal stages and the three-stage decomposition provides complementary, stage-aware coverage.

\section{Additional Analyses}

\subsection{Charge-Level Effects}
\label{app:charge}

Fig.~\ref{fig:charge} counts, for the ten most frequent charges in the CAIL-2018 test split, the test predictions that editing flips. On both backbones, flips from incorrect to correct outnumber flips from correct to incorrect for every one of these charges, so the gain is not concentrated in a few charges.

\subsection{Pair-Level Margin Changes}
\label{app:quadrant}

Fig.~\ref{fig:shift_scatter} plots the margin change of both members of every strict false-guilty test pair under PROSE. A uniform output shift would place all pairs at one point on the diagonal; instead, the displacements spread widely across pairs, and within a pair the charge-not-established member tends to move slightly further than its counterpart. The median changes $\Delta s_{NG}$ and $\Delta s_{G}$ are $-0.51$ and $-0.47$ with Qwen and $-0.19$ and $-0.13$ with GLM, and the median within-pair difference is $-0.066$ with Qwen (95\% CI $[-0.106,-0.031]$) and $-0.049$ with GLM (95\% CI $[-0.062,-0.041]$). The advantage over a uniform shift therefore comes mainly from displacements that vary across pairs. Fig.~\ref{fig:quadrant} shows the underlying margins of both members of every test pair before and after editing.



\end{document}